\documentclass[10pt,twocolumn,letterpaper]{article}

\usepackage[pagenumbers]{cvpr}      

\usepackage{listings}
\lstdefinestyle{prompt}{
  basicstyle=\ttfamily\scriptsize,
  breaklines=true,
  breakatwhitespace=true,
  breakindent=1em,
  columns=fullflexible,
  keepspaces=true,
  frame=single,
  framerule=0.4pt,
  rulecolor=\color[gray]{0.6},
  xleftmargin=2pt,
  xrightmargin=2pt,
  aboveskip=6pt,
  belowskip=6pt,
  showstringspaces=false,
}

\usepackage{booktabs}
\usepackage{array}
\usepackage{tcolorbox}

\tcbuselibrary{listings,breakable}
\usepackage{capt-of}
\definecolor{promptgold}{HTML}{7197B0}
\definecolor{promptcream}{HTML}{F3F7FA}
\newcommand{\egseprompt}[4][black]{%
  \begin{figure*}[p]
  \tcbinputlisting{listing file={#3},listing only,
    title={#2},fonttitle=\sffamily\bfseries\small,coltitle=white,
    colback=promptcream,colbacktitle=promptgold,colframe=promptgold,
    boxrule=0.8pt,arc=3mm,left=3mm,right=3mm,top=2mm,bottom=2mm,
    listing options={style=prompt,basicstyle=\color{#1}\sffamily\fontsize{8}{8.5}\selectfont,
      frame=none,xleftmargin=0pt,xrightmargin=0pt,aboveskip=0pt,belowskip=0pt}}
  \caption{#4}
  \end{figure*}
}

\AtBeginDocument{\hypersetup{hypertexnames=false}}

\definecolor{cvprblue}{rgb}{0.21,0.49,0.74}
\usepackage[pagebackref,breaklinks,colorlinks,allcolors=cvprblue]{hyperref}

\def\confName{CVPR}
\def\confYear{2027}

\title{EGSE: Evidence-Grounded Skill Evolution for Medical Reasoning Agents}

\author{%
Mingyuan Li$^{1,2}$, Tong Jia$^{1}$, Ziyuan Yang$^{2}$, Wenbo Qi$^{2}$,\\
Haoran Wang$^{1}$, Jie Gui$^{3}$, Baosheng Yu$^{2}$\\[4pt]
{\small $^{1}$College of Information Science and Engineering, Northeastern University, China}\\
{\small $^{2}$Nanyang Technological University\qquad $^{3}$Southeast University}\\[4pt]
{\small Tong Jia: \href{mailto:jiatong@ise.neu.edu.cn}{\texttt{jiatong@ise.neu.edu.cn}}\qquad
Baosheng Yu: \href{mailto:baosheng.yu@ntu.edu.sg}{\texttt{baosheng.yu@ntu.edu.sg}}}
}

\newlength{\EGSEOriginalCaptionSkip}
\newlength{\EGSEOriginalTextFloatSep}
\newlength{\EGSEOriginalDblTextFloatSep}
\newlength{\EGSEOriginalFloatSep}
\newlength{\EGSEOriginalDblFloatSep}
\newlength{\EGSEOriginalInTextSep}
\AtBeginDocument{%
  \setlength{\EGSEOriginalCaptionSkip}{\abovecaptionskip}%
  \setlength{\EGSEOriginalTextFloatSep}{\textfloatsep}%
  \setlength{\EGSEOriginalDblTextFloatSep}{\dbltextfloatsep}%
  \setlength{\EGSEOriginalFloatSep}{\floatsep}%
  \setlength{\EGSEOriginalDblFloatSep}{\dblfloatsep}%
  \setlength{\EGSEOriginalInTextSep}{\intextsep}%
  \captionsetup[figure]{skip=4pt}
  \captionsetup[table]{skip=4pt}
  \setlength{\textfloatsep}{6pt plus 1pt minus 1pt}
  \setlength{\dbltextfloatsep}{6pt plus 1pt minus 1pt}
  \setlength{\floatsep}{6pt plus 1pt minus 1pt}
  \setlength{\dblfloatsep}{6pt plus 1pt minus 1pt}
  \setlength{\intextsep}{6pt plus 1pt minus 1pt}
}
\hypersetup{
  pdftitle={EGSE: Evidence-Grounded Skill Evolution for Medical Reasoning Agents},
  pdfauthor={Mingyuan Li, Tong Jia, Ziyuan Yang, Wenbo Qi, Haoran Wang, Jie Gui, Baosheng Yu}
}
\begin{document}
\maketitle
\begin{abstract}
Medical reasoning agents combine foundation models with specialized tools to interpret medical images, yet reusing their historical experience remains challenging. A correct final answer can conceal flawed reasoning or tool use, synthesized advice may lack source support, and superficially similar cases may require different diagnostic procedures. In this paper, we propose EGSE, an evidence-grounded skill evolution framework that distills historical dialogues from similar cases into reusable conditional workflows and progressively refines a multi-faceted skill library through supported and validated updates to improve medical agents. EGSE evaluates intermediate reasoning and tool use and contrasts repeated successful and failed rollouts to extract useful practices and evidence-supported avoidance guidance. These experiences are synthesized into workflows with applicability conditions and checks traceable to source dialogues. Library additions and merges require source-grounded review and execution validation. At inference time, EGSE routes queries to skill nodes organized by imaging modality, question task type, and disease category, then retrieves and selects applicable workflows based on their conditions, the current images, and the question to guide evidence acquisition and interpretation. Experiments across four chest X-ray benchmarks show that EGSE outperforms representative medical agents and skill evolution methods in reasoning accuracy while reducing tool calls. It also delivers consistent accuracy gains across seven foundation models and supports skill transfer across datasets.
\end{abstract}

\section{Introduction}
\label{sec:intro}

Medical image interpretation requires integrating visual findings with clinical context and selecting appropriate procedures to distinguish competing explanations. Building on medical foundation models~\cite{li2023llava,moor2023med,chen2024chexagent,bannur2024maira}, medical agents~\cite{fallahpour2025medrax,lou2026cxragent} integrate LLM~\cite{bai2025qwen3,hurst2024gpt} reasoning with specialized neural tools~\cite{cohen2022torchxrayvision,lian2021structure} to acquire visual evidence through classification, detection, segmentation, and visual question answering. For example, MedRAX~\cite{fallahpour2025medrax} integrates chest X-ray analysis tools into interactive reasoning, while CXRAgent~\cite{lou2026cxragent} combines tool-output validation with diagnostic planning and expert collaboration. These advances support evidence acquisition within individual cases~\cite{li2024mmedagent,zhou2025mam,wang2026medagent}. Across cases posing similar clinical questions, however, the logic of what evidence to acquire and how to interpret it can be reused~\cite{sun2026experience,ni2026trace2skill,zhao2024expel}. Extracting this logic from historical rollouts and invoking it appropriately in new cases remain challenging.

\begin{figure*}[t]
  \centering
  \includegraphics[width=\textwidth]{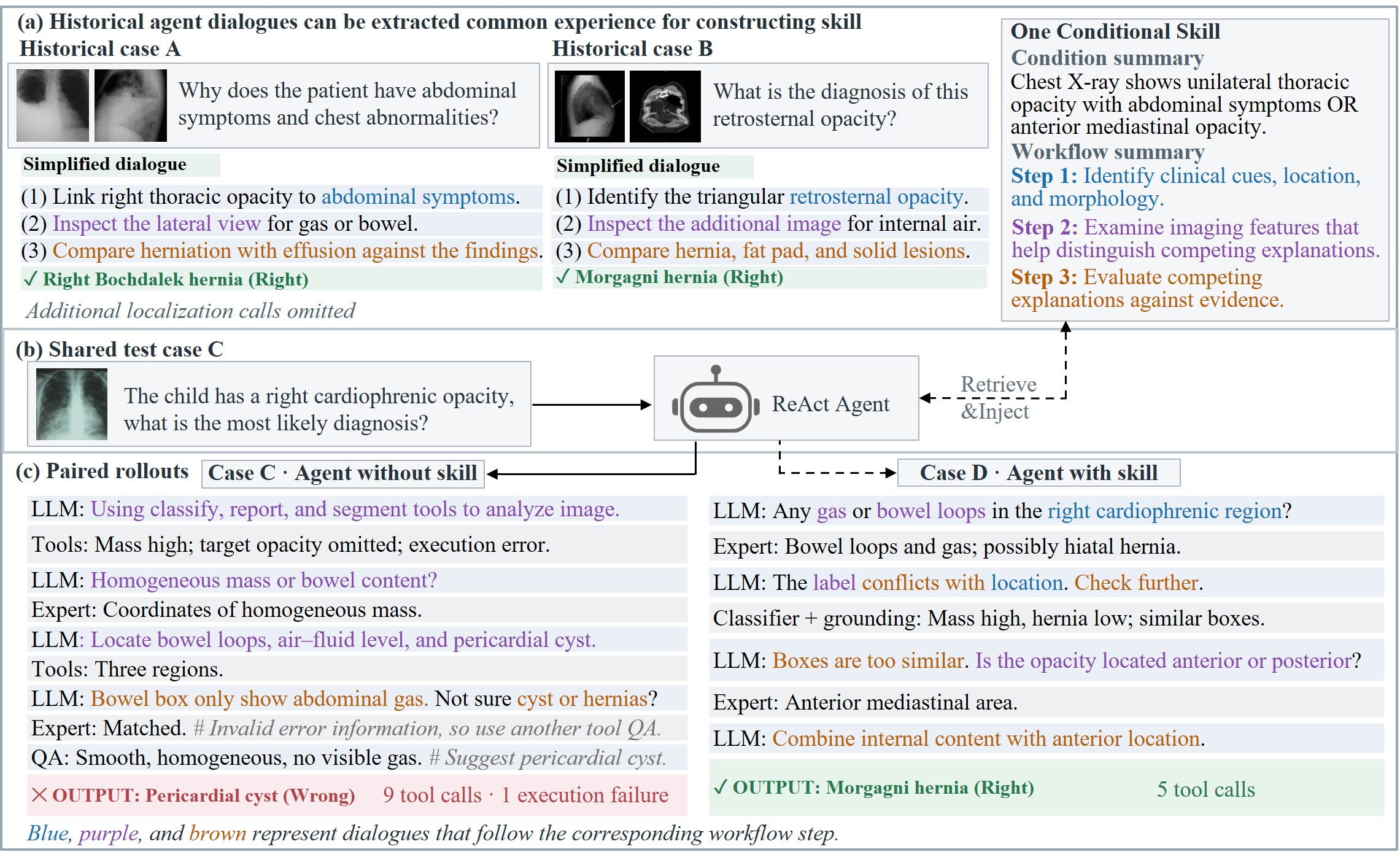}
  \caption{Illustration of conditional skills distilled from historical agent dialogues for medical image interpretation. A conditional skill abstracts reusable reasoning patterns into three stages: identifying relevant cues, examining discriminative imaging features, and evaluating competing explanations. \textbf{Without} the skill, the ReAct agent blindly invokes tools, pursues misleading intermediate findings, and overlooks question-specific clinical cues, resulting in an incorrect diagnosis of pericardial cyst after nine tool calls and one execution failure.  \textbf{With} the skill, the agent integrates imaging findings with anatomical context and correctly identifies a Morgagni hernia in only five tool calls.}
  \label{fig:figure1}
\end{figure*}

Recent medical-agent research explores experience-driven self-evolution through memory-based reflection in AMC~\cite{lan2024depression}, composite-tool discovery in MACRO~\cite{fan2026evolving}, template and tool evolution in STELLA~\cite{jin2025stella}, experience-guided planning in HealthFlow~\cite{zhu2025healthflow}, and skill-memory evolution in SkeMeX~\cite{sun2026experience}. Across these approaches, historical interactions provide feedback for extracting reusable procedural knowledge, which is stored as memories, tools, plans, or skills and retrieved or invoked to guide subsequent tasks. Turning this experience into reliable workflows raises three questions. \emph{First, are the extracted practices and accepted updates useful?} Final outcomes alone do not identify useful intermediate practices, extracted advice can remain specific to its source case, and an update can impair an existing workflow. \emph{Second, is the synthesized workflow supported?} Even useful-looking content can introduce advice absent from the source dialogues. \emph{Third, is a retrieved workflow applicable to the current case?} Superficially similar cases may differ in anatomy, visual findings, or question requirements. Figure~\ref{fig:figure1} illustrates the opportunity: historical cases share a procedure for identifying relevant cues, examining discriminative imaging features, and evaluating competing explanations. Applied to a new case, this procedure guides the agent to verify anatomical location and internal contents before resolving competing diagnoses. This motivates our hypothesis that useful, source-supported practices can be distilled into conditional workflows and reused when their conditions match new cases.

To realize this hypothesis, we propose EGSE, an evidence-grounded skill evolution framework that progressively refines an external multi-faceted skill library for medical image interpretation. Each skill node is indexed by imaging modality, question task type, and disease category and contains multiple conditional workflows. To identify useful practices, EGSE scores intermediate reasoning and tool use against ground-truth answers, extracts candidate recommendations from positively scored content shared by successful rollouts and avoidance advice from negatively scored content shared by failed rollouts, and uses semantic subtraction to retain recommendations that do not conflict with the avoidance advice under overlapping applicability conditions. These operations aim to retain reusable workflow steps, termed checks to distinguish them from dialogue steps, rather than case-specific conclusions. Execution validation tests proposed changes on the current case and a previous case belonging to the same skill node to screen for harmful updates. To ground synthesis in source evidence, each workflow check cites atomic experiences linked to source dialogue-step pairs, and source-based review assesses these support links. To assess applicability, multi-faceted routing identifies a relevant node, while condition-based retrieval and image-aware selection choose among its workflows. Only changes that pass review and execution validation enter the library; both model parameters and the library remain frozen at test time.

We evaluate EGSE on four chest X-ray benchmarks, including ChestAgentBench~\cite{fallahpour2025medrax}, CheXbench~\cite{chen2024chexagent}, MIMIC-CXR~\cite{johnson2019mimic}, and SLAKE-VQA~\cite{liu2021slake}. On ChestAgentBench, with Qwen3.7-Flash~\cite{qwen37} as the foundation model, EGSE improves accuracy from 73.75\% to 77.70\% over MedRAX while reducing average tool calls per case from 5.8 to 2.3. It also outperforms the general-domain skill-evolution methods Trace2Skill~\cite{ni2026trace2skill} and MACRO, as well as the medical-domain method SkeMeX. 
Across seven foundation models, EGSE consistently improves accuracy on ChestAgentBench, with gains ranging from 2.58\% to 5.84\%. We further evaluate cross-dataset skill transfer and within-dataset skill construction on external datasets. On CheXbench, the frozen skill library constructed on ChestAgentBench improves accuracy by 1.7\%, while a library constructed from CheXbench's own construction split yields a 3.3\% accuracy gain.
Our main contributions are summarized as follows:
\begin{itemize}
    \item We propose EGSE, an evidence-grounded skill evolution framework that distills historical medical-agent dialogues into reusable conditional workflows without updating model parameters.
    \item We integrate quality-aware contrastive experience extraction, source-grounded synthesis, and execution validation to extract useful practices and validate skill updates, and organize the resulting skills in a multi-faceted skill library for condition-aware reuse.
    \item We demonstrate that EGSE improves medical reasoning accuracy while reducing tool calls, with consistent gains across seven foundation models and further improvements through both cross-dataset transfer and within-dataset skill construction.
\end{itemize}

\section{Related Work}
\label{sec:related}

\noindent\textbf{Medical Agents.}
Medical vision--language models~\cite{singhal2023large,chen2023meditron,li2023llava,moor2023med,chen2024vision} provide a foundation for image-based clinical reasoning. For example, LLaVA-Med~\cite{li2023llava} adapts LLaVA~\cite{liu2023visual} to the biomedical domain through instruction tuning, while CheXagent~\cite{chen2024chexagent} specializes in chest X-ray interpretation.
Beyond model adaptation, MedRAG~\cite{xiong2024benchmarking} and i-MedRAG~\cite{xiong2024improving} introduce retrieval and iterative information gathering, while MMedAgent~\cite{li2024mmedagent} supports medical tool use. MedAgents~\cite{tang2024medagents}, MDAgents~\cite{kim2024mdagents}, MedAgent-Pro~\cite{wang2026medagent} explore expert collaboration, adaptive coordination, and clinically guided planning.
For chest X-ray reasoning, MedRAX~\cite{fallahpour2025medrax} couples a foundation model with specialized medical tools~\cite{bannur2024maira,chambon2024chexpert,chen2024chexagent,cohen2022torchxrayvision,li2023llavamed,lian2021structure} in an ReAct-style~\cite{yao2022react} loop, providing a baseline for our study. CXRAgent~\cite{lou2026cxragent} uses a central director to coordinate tool invocation, visual-evidence-based output validation, diagnostic planning, and expert collaboration. These approaches improve evidence acquisition and coordination within a case, while learning reusable experience across historical cases introduces a complementary direction for medical agents.

\noindent\textbf{Self-Evolving Agents.}
Reflection and experience abstraction enable agents to improve subsequent decisions from interaction feedback~\cite{shinn2023reflexion,zhao2024expel}. Executions can be distilled into reusable code skills, workflows, and verifiable functions~\cite{wang2023voyager,wang2025agent,wang2024trove}, while trainable memory and skill-augmented reinforcement learning provide parameter-learning routes to adaptation~\cite{zhang2026memskill,xia2026skillrl}.
Trace2Skill~\cite{ni2026trace2skill} extracts trajectory-local lessons from successful and failed executions and consolidates them in parallel into portable procedural skills, supporting reuse across models without parameter updates.
SkillComposer~\cite{zhang2026skillcomposer} balances specificity and generalization through skill creation, specialization, and merging; GRASP~\cite{moll2026grasp} uses probes to validate proposed updates. Evo-Memory~\cite{wei2025evo} further integrates memory refinement with reasoning and action.

In medicine, recent work develops skills from both clinical knowledge and interaction experience.
GuideSkill~\cite{cao2026guideskill} converts human-designed clinical guidelines into executable rules for improving reasoning performance.
MACRO~\cite{fan2026evolving} converts recurring effective tool-call sequences from verified medical trajectories into reusable composite tools. It uses image-feature memory to inform tool selection and trains the agent to invoke the discovered composites more reliably.
SkeMeX~\cite{sun2026experience} distills interaction trajectories into reusable skills and uses their performance on subsequent tasks to guide skill retrieval and retention in an external repository, without updating model weights. 
However, extracting useful experience from historical dialogues, grounding synthesized advice in the source dialogues, and restricting skill reuse to related cases that call for the same reasoning procedures remain insufficiently addressed.

\section{Method}
\label{sec:method}

\begin{figure*}[t]
  \centering
  \includegraphics[width=\textwidth]{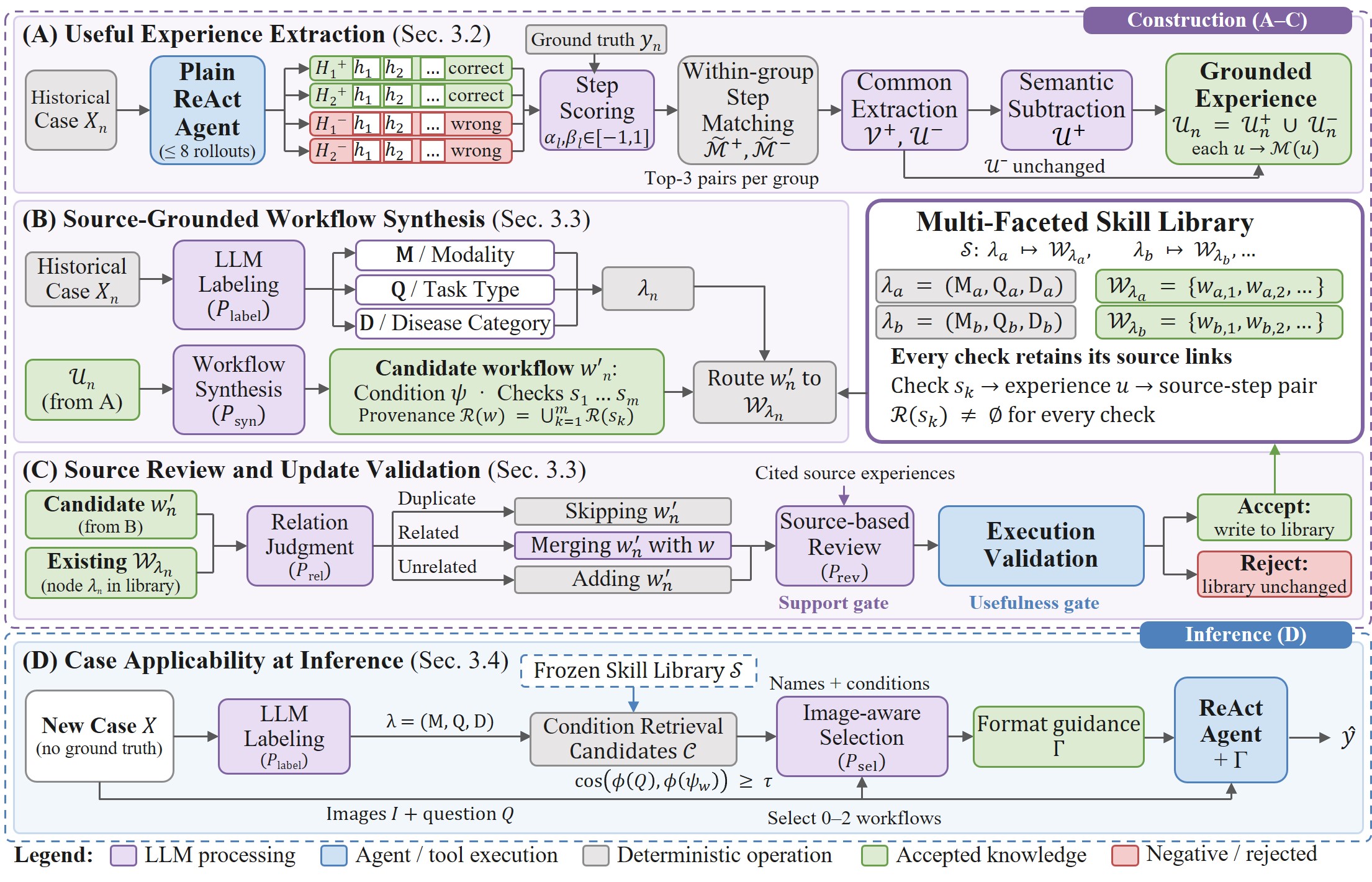}
  \caption{Overview of EGSE. \textbf{(A)} Step scoring and contrastive extraction identify useful practices, while semantic subtraction filters conflicting recommendations. \textbf{(B)} Source-linked experiences are synthesized into conditional workflows and routed to the multi-faceted skill library. \textbf{(C)} Proposed additions and merges enter the library only after source-based review and execution validation. \textbf{(D)} Multi-faceted routing, condition retrieval, and image-aware selection provide applicable workflows for new cases with the library frozen.}
  \label{fig:overview}
\end{figure*}

EGSE distills historical medical-agent dialogues into conditional workflows stored in an external multi-faceted skill library (Figure~\ref{fig:overview}). It addresses the three questions in Section~\ref{sec:intro} through step scoring, contrastive extraction, and execution validation for usefulness, source-linked synthesis and review for content support, and condition-aware retrieval and selection for applicability. Both the agent parameters and the constructed library remain frozen at test time.

\subsection{Problem Formulation}
\label{sec:problem}

We consider a ReAct-style medical agent equipped with a foundation model $f_\theta$ and a set of neural tools $\mathcal{T}$. Its input is a case $X=(I,Q)$, where $I$ denotes the medical image(s) and $Q$ the clinical question, including any answer options. At step $l$, the agent generates a reasoning message $C_l$ and a set of tool calls $\mathcal{A}_l$, conditioned on its previous interaction history $H_{<l}$ and external workflow guidance $\Gamma$:
\begin{equation}
\label{eq:react}
(C_l,\mathcal{A}_l)=f_\theta(X,H_{<l};\Gamma).
\end{equation}
Each call specifies a tool in $\mathcal{T}$ and its arguments. The returned outputs form the evidence $E_l$, and the step $(C_l,\mathcal{A}_l,E_l)$ is appended to the history. A rollout produces a dialogue $H$ and a final answer $\hat y$; it ends when no tool is requested or the step budget is reached, in which case the agent is instructed to answer directly. For a skill-retrieved agent, the content quality of $\Gamma$ affects the fianl result. The plain agent uses $\Gamma=\emptyset$.

Given construction cases $\mathcal{D}_{\mathrm{build}}=\{(X_n,y_n)\}_{n=1}^{N}$ with ground-truth answers, where $X_n=(I_n,Q_n)$, our goal is to distill historical agent dialogues into an external skill library $\mathcal{S}$ that improves reasoning performance on unseen cases without updating $\theta$. For each new case $X$, selected workflows from $\mathcal{S}$ are formatted as guidance $\Gamma$ for the agent. Ground-truth answers are available during library construction, whereas test-time routing and selection use only $X$. Both the model parameters and the constructed library remain frozen during testing.

\subsection{Extracting Useful Experience}
\label{sec:experience}

A correct final answer does not certify the reasoning and tool use that produced it. Conversely, an unsuccessful rollout may contain valid intermediate steps. As shown in Figure~\ref{fig:overview}(a), EGSE therefore assesses step quality before extracting practices shared across repeated rollouts, then uses observed failures to filter conflicting recommendations.

\noindent\textbf{Contrastive rollouts and step scoring.}
For a fixed construction case $(X_n,y_n)$, we omit the case index $n$ in this subsection for readability. We sample the plain agent until two successful dialogues $H_1^+,H_2^+$ and two unsuccessful dialogues $H_1^-,H_2^-$ are collected. Cases that do not yield both groups within eight rollouts are skipped. 

We denote by $f_{\mathrm{ext}}$ the same foundation model as $f_\theta$, invoked with task-specific prompts in separate stateless contexts to keep experience-processing context separate from image-interpretation context. For each dialogue, $f_{\mathrm{ext}}$ independently assigns each step $l$ a reasoning score $\alpha_l$ and a tool-use score $\beta_l$, both in $[-1,1]$, using the scoring prompt $P_{\mathrm{score}}$ provided in Appendix~\ref{supp:prompt_score}:
\begin{equation}
\label{eq:score}
\bigl((\alpha_l,\beta_l,r_l^{\alpha},r_l^{\beta})\bigr)_{l=1}^{L_H}
=f_{\mathrm{ext}}(X,H,y;P_{\mathrm{score}}).
\end{equation}
Here, $L_H$ is the number of steps in dialogue $H$. The reasoning score $\alpha_l$ assesses the validity and contribution of $C_l$ toward the ground-truth answer $y$ in the context of the case input $X$ and the full dialogue $H$. The tool-use score $\beta_l$ assesses the tool calls $\mathcal{A}_l$ together with their returned evidence $E_l$. The rationales $r_l^{\alpha}$ and $r_l^{\beta}$ explain their corresponding scores based on the dialogue content.

\noindent\textbf{Score-guided common extraction.}
Within each rollout group, we compute similarities between steps from the two dialogues using a text encoder $\phi$, as bge-base-en-v1.5~\cite{xiao2024c}. When both steps include tool calls, their similarity is the average of the cosine similarities between their reasoning texts, tool-call texts, and returned-evidence texts. When either or both steps contain no tool calls, their similarity is computed from reasoning texts alone. We retain the three most similar step pairs in each group, denoted by $\mathcal{M}^+$ and $\mathcal{M}^-$. We denote the annotated pair sets by $\widetilde{\mathcal{M}}^+$ and $\widetilde{\mathcal{M}}^-$: Each pair retains its source identifiers and compared components and is annotated with the scores and assessment rationales \((\alpha_l,\beta_l,r_l^\alpha,r_l^\beta)\) for each of its two source steps.

The common-experience prompt $P_{\mathrm{com}}$, as shown in Appendix~\ref{supp:prompt_common}, extracts positive candidates $\mathcal{V}^+$ from the successful group and avoidance advice $\mathcal{U}^-$ from the unsuccessful group:
\begin{equation}
\label{eq:common}
\begin{aligned}
\mathcal{V}^+ &= f_{\mathrm{ext}}(\widetilde{\mathcal{M}}^+,Q;P_{\mathrm{com}}),\\
\mathcal{U}^- &= f_{\mathrm{ext}}(\widetilde{\mathcal{M}}^-,Q;P_{\mathrm{com}}).
\end{aligned}
\end{equation}
A successful rollout may contain flawed steps, while a failed rollout may contain useful ones. Therefore, within each matched pair, reasoning content is eligible for positive or negative extraction only when both source steps have positive or negative \(\alpha_l\) scores, respectively. The same criterion applies to tool-use content using $\beta_l$. The corresponding rationales $r_l^\alpha$ or $r_l^\beta$ and the original steps must support the extracted recommendation or avoidance advice.

\noindent\textbf{Semantic subtraction.}
Recommendations extracted from successful rollouts may still conflict with avoidance advice extracted from failed rollouts. The positive-experience prompt $P_{\mathrm{pos}}$ (Appendix~\ref{supp:prompt_pos}) therefore applies semantic subtraction to filter conflicting recommendations and obtain retained recommendations $\mathcal{U}^+$:
\begin{equation}
\label{eq:positive}
\mathcal{U}^+=f_{\mathrm{ext}}(\mathcal{V}^+,\mathcal{U}^-;P_{\mathrm{pos}}).
\end{equation}
It compares the underlying behaviors and applicability conditions, removing recommendations targeted by avoidance advice under overlapping conditions. A candidate is dropped when conflict cannot be ruled out. A partial remainder is retained only if its original positive sources independently support it, and $\mathcal{U}^-$ remains unchanged. 

The retained recommendations and avoidance advice form the experience set $\mathcal{U}_n=\mathcal{U}_n^+\cup\mathcal{U}_n^-$ for case $n$. Each experience $u\in\mathcal{U}_n$ records its advice, applicability condition, evidence notes, and limitations. It also stores references to its supporting dialogue-step pairs, whose set is denoted by $\mathcal{M}(u)\neq\emptyset$. Each pair retains the identifiers of its two source steps, allowing the experience to be traced back to the original dialogues. These references allow each experience to be traced back to the original dialogue steps.

\subsection{Source-Grounded Workflow Construction and Update}
\label{sec:skill}
\label{sec:update}

Useful experience must be converted into reusable procedures without introducing unsupported advice. As shown in Figure~\ref{fig:overview}(b), EGSE synthesizes workflows from the extracted records and preserves the source links of each check. Proposed updates undergo source-based review and execution validation (Figure~\ref{fig:overview}(c)). The former assesses whether the cited sources support the updates, while the latter tests whether they help on the current case without failing a recent case in the same skill node.

\noindent\textbf{Library organization.}
The library groups workflows by three facets: imaging modality (M), question task type (Q), and disease category of the question topic (D). These facets describe the image source, the requested operation, and the clinical topic, respectively. They contain 13, 9, and 29 labels, each including an unknown label; disease categories follow the chapters of ICD-11~\cite{who2026icd11}. The labeling prompt $P_{\mathrm{label}}$ assigns a facet tuple $\lambda$ from the case alone:
\begin{equation}
\label{eq:mqd}
\lambda(X)=f_\theta(X;P_{\mathrm{label}}).
\end{equation}
For construction case $X_n$, we write $\lambda_n=\lambda(X_n)$. Each tuple addresses one skill node, whose workflows form a set $\mathcal{W}_\lambda$. Facets provide a coarse relevance filter, while individual workflow conditions distinguish situations within a node. Label definitions and routing rules are given in Appendices~\ref{supp:prompt_label} and~\ref{supp:mqd}.

\noindent\textbf{Workflow synthesis with source links.}
A workflow $w$ consists of a short name, an applicability condition $\psi$, an ordered sequence of workflow steps, termed checks $s_1,\ldots,s_m$, and provenance $\mathcal{R}(w)$. Each check specifies what to determine, how to acquire or examine case evidence, and how the findings, including unresolved findings, guide the next decision. Here, case evidence refers to imaging findings and tool outputs for the case being analyzed.

For each construction case with nonempty $\mathcal{U}_n$, the synthesis prompt $P_{\mathrm{syn}}$ produces at most one candidate workflow $w'_n=f_{\mathrm{ext}}(\mathcal{U}_n;P_{\mathrm{syn}})$ (Appendix~\ref{supp:prompt_syn}). Each check $s_k$ cites a nonempty set of supporting experiences $\mathcal{R}(s_k)\subseteq\mathcal{U}_n$ and must preserve their conditions, advice, and limitations. The workflow provenance is $\mathcal{R}(w)=\bigcup_{k=1}^{m}\mathcal{R}(s_k)$. Each check is linked to its supporting experiences and their source steps:
\begin{equation}
\label{eq:provenance}
s_k \to u\in\mathcal{R}(s_k)
\to (h_a,h_b)\in\mathcal{M}(u).
\end{equation}
The arrows denote stored reference links, and $h_a$ and $h_b$ are source steps from two historical dialogues. These links are preserved during merging and used to review whether the cited experiences support each check. The candidate $w'_n$ is assigned to the skill node indexed by $\lambda_n$.

\noindent\textbf{Adding, merging, or skipping.}
Candidates are processed in increasing case-index order. The relation prompt $P_{\mathrm{rel}}$ compares $w'_n$ with the existing workflows $w\in\mathcal{W}_{\lambda_n}$ by check content and applicability conditions (Appendix~\ref{supp:prompt_rel}). It judges a pair as \textsc{duplicate} when the existing workflow covers every candidate check, how its results are used, and its applicability condition; \textsc{related} when the workflows examine the same findings for the same competing explanations and the candidate supplies a source-supported refinement or applicability extension; and \textsc{unrelated} otherwise.


For each \textsc{related} pair, $P_{\mathrm{rel}}$ also specifies the applicability-condition adjustment required for merging. With \emph{none}, the existing condition is retained and only the checks are refined. With \emph{conditional}, the existing condition is retained, and the candidate's checks apply only when its narrower condition is met. With \emph{generalize}, the condition is broadened to cover both situations, while checks specific to either situation remain conditional on it.

A candidate is skipped if any existing workflow is classified as \textsc{duplicate}. Otherwise, EGSE selects a merge target from the \textsc{related} workflows, prioritizing \emph{none}, then \emph{conditional}, then \emph{generalize}. The merge prompt $P_{\mathrm{merge}}$ then combines the selected workflow and the candidate (Appendix~\ref{supp:prompt_merge}). Every change to existing content must be grounded in the candidate's cited experiences, and retained checks preserve their source links. If the candidate is classified as \textsc{unrelated} to all existing workflows in the node, it is proposed as a new workflow. If the node contains no workflows, the candidate is directly proposed as its first workflow. In both cases, the addition requires source-based review and execution validation before acceptance.

\noindent\textbf{Source-based review.}
After output format and reference validity are checked by code, $P_{\mathrm{rev}}$ verifies whether each check \(s\) in the proposed workflowis supported by its source dialogue-step pairs, traced through \(\mathcal{R}(s)\) and \(\mathcal{M}(u)\) (Appendix~\ref{supp:prompt_rev}). The review identifies unsupported instructions and applicability conditions that extend beyond the source support. A failing workflow is regenerated at most twice using the reported violations and is discarded if it still fails.

\noindent\textbf{Execution validation.}
Execution validation checks whether a proposed library update harms performance on the current case or a recent case previously routed to the same skill node. Let \(\mathcal{S}'\) denote the library obtained by provisionally applying the reviewed addition or merge to \(\mathcal{S}\). The validation cases comprise \((X_n,y_n)\) and, for a nonempty node, \((X_j,y_j)\), where \(j<n\) is the largest index among earlier cases routed to the same node (\(\lambda_j=\lambda_n\)). Each case is answered three times using \(\mathcal{S}'\) and the inference procedure in Section~\ref{sec:inference}, and we commit \(\mathcal{S}\leftarrow\mathcal{S}'\) only if every answer passes the task-specific success criterion. Otherwise, \(\mathcal{S}\) remains unchanged.
\subsection{Selecting Applicable Workflows at Inference}
\label{sec:inference}

A source-supported workflow can still be inappropriate for a new case. EGSE therefore combines multi-faceted routing, condition-based retrieval, and image-aware selection. Both the model parameters and the skill library remain frozen during test inference.

First, the labeling procedure in Eq.~\eqref{eq:mqd} routes $X$ to the node indexed by $\lambda=\lambda(X)$. Let $\mathcal{C}$ denote the candidate workflows retrieved from that node, and $\tau=0.67$ the condition-similarity threshold. Retrieval compares the question with each workflow's applicability condition:
\begin{equation}
\label{eq:retrieval}
\mathcal{C}=\{w\in\mathcal{W}_\lambda\mid
\cos(\phi(Q),\phi(\psi_w))\ge\tau\},
\end{equation}
where $\psi_w$ is the condition of workflow $w$. The same threshold is used during construction-time execution validation.

Text similarity cannot establish whether a condition agrees with the visual evidence. The selection prompt $P_{\mathrm{sel}}$ therefore receives the current images and question together with each candidate's name and condition, without its checks or source references (Appendix~\ref{supp:prompt_sel}). It selects workflow identifiers by applicability without first guessing the diagnosis. The selected workflows' names, conditions, and checks are formatted as guidance $\Gamma$ for the agent:
\begin{equation}
\label{eq:selection}
\Gamma=\operatorname{Format}\!\left(f_\theta(X,\mathcal{C};P_{\mathrm{sel}})\right),
\end{equation}
\(\operatorname{Format}\) formats the names, conditions, and checks of the selected workflows as guidance text. Selection permits up to two workflows addressing different evidence tasks, such as localization and differential reasoning, or none when no candidate applies. This limit controls context length to reduce degradation in reasoning caused by excessive context. Finally, $\Gamma$ is injected into the agent through Eq.~\eqref{eq:react}; an empty selection recovers the plain agent.

\section{Experiments}
\label{sec:experiments}

We evaluate whether EGSE improves accuracy and tool-use efficiency, generalizes across foundation models and datasets. We further analyze how its components affect workflow usefulness, groundedness, and applicability.

\subsection{Experimental Setup}
\label{sec:exp_setup}

\noindent\textbf{Datasets and metrics.}
We evaluate on four chest X-ray benchmarks. ChestAgentBench~\cite{fallahpour2025medrax} contains 2,500 clinical cases from 609 patients, partitioned into 1,498/503/499 construction/development/test cases from 365/122/122 patients without patient overlap. We report overall and per-task accuracy. CheXbench~\cite{chen2024chexagent} covers visual question answering and image--text reasoning, evaluated by source-subset and pooled accuracy. SLAKE-VQA~\cite{liu2021slake} uses exact-match accuracy and word recall. MIMIC-CXR~\cite{johnson2019mimic} evaluates report generation using CheXbert-based~\cite{smit2020combining} micro-/macro-F1 over 14 observations and five findings, following M4CXR~\cite{park2025m4cxr}. All performance metrics are percentages; pool definitions and metric details appear in Appendix~\ref{supp:exp_protocol}.

\noindent\textbf{Evaluation settings.}
On ChestAgentBench, skills are constructed on the construction partition and configurations are selected on the development partition. For external evaluation, cross-dataset transfer applies the ChestAgentBench library to each full target evaluation pool; dataset generalization builds a separate library using a grouped 4:1 construction/test split within each target and evaluates on its test portion. All libraries remain frozen during testing. Appendix~\ref{supp:dataset_specific} details data access and grouping.

\noindent\textbf{Comparisons.}
We compare with LLaVA-Med~\cite{li2023llava}, CheXagent~\cite{chen2024chexagent}, Direct FM, MedRAX~\cite{fallahpour2025medrax}, CXRAgent~\cite{lou2026cxragent}, Trace2Skill~\cite{ni2026trace2skill}, SkeMeX~\cite{sun2026experience}, and MACRO~\cite{fan2026evolving}. MACRO is adapted without parameter training, retaining its memory and composite-tool mechanisms. MedRAX and its skill variants share the FM, underlying tools, partitions, and inference limits; skill methods receive the same construction-data access.

\noindent\textbf{Implementation.}
Qwen3.7-Flash is the default backbone and extraction model, with rollout cap $R=8$ and reasoning-step budget $L=7$. We evaluate seven FMs from the Qwen and GPT families. Decoding, embeddings, hardware, and backbone settings are detailed in Appendix~\ref{supp:exp_models}.


\begin{table}[!ht]
\centering
\caption{Model performance on ChestAgentBench.}
\label{tab:main_results}
\setlength{\tabcolsep}{2pt}
\begin{tabular*}{\linewidth}{@{\extracolsep{\fill}}lrrrr@{}}
\toprule
Method & Accuracy$\uparrow$ & Calls$\downarrow$ & Rounds$\downarrow$ & Score$\uparrow$ \\
\midrule
LLaVA-Med &28.70& --- & 1 & --- \\
CheXagent & 39.50 & --- & 1 & --- \\
\hline
Direct FM & 70.80 & --- & 1 & --- \\
+MedRAX & 73.75 &  5.8 & 3.9 & 0.14 \\
+Trace2Skill & 74.83 & 6.1 & 4.3 & 0.13 \\
+MACRO$^*$ & 76.31& 4.9&4.5&0.19\\
+CXRAgent & 76.53& 7.1&11.3&0.12\\
+SkeMeX & 76.87 & 3.1& 2.9 & 0.26 \\
+EGSE & \textbf{77.70} & \textbf{2.3} & \textbf{2.6} & \textbf{0.31} \\
\bottomrule
\end{tabular*}\par\vspace{2pt}
{\raggedright\fontsize{7}{8}\selectfont Accuracy is reported in percent; Calls and Rounds are averages per case. {Score averages tool-bearing steps using Eq.~\eqref{eq:score}.} Dashes indicate inapplicable tool metrics for direct prediction. Matched agent variants use the same underlying FM, tools, and data partitions. MACRO$^*$ retains memory and composite-tool discovery without parameter updates. \textbf{Bold}: best agent results.\par}
\end{table}

\vspace{-10pt}
\subsection{{Main Results}}

\vspace{-2pt}

\noindent\textbf{Overall comparison.}
Table~\ref{tab:main_results} compares EGSE with medical foundation models, tool-based agents, and skill-evolution methods on ChestAgentBench. EGSE achieves the highest accuracy of 77.70\%, exceeding Direct FM (Qwen3.7-Flash) and MedRAX by 6.90\% and 3.95\%, respectively, and the strongest skill-evolution baseline, SkeMeX, by 0.83\%. Compared with MedRAX, it reduces tool calls by 60.3\% (5.8 to 2.3) and reasoning rounds from 3.9 to 2.6, while increasing the tool-step score from 0.14 to 0.31. These results show that EGSE outperforms the evaluated medical skill-based self-evolution methods in accuracy while using fewer tool calls and achieving a higher tool-step score.

\subsection{{Generalization across FMs and Datasets}}

\begin{table*}[!ht]
\centering
\caption{Evaluation of EGSE across foundation models and through cross-dataset skill transfer and within-dataset skill construction.}
\label{tab:generalization}
\label{tab:transfer}
\setlength{\tabcolsep}{3pt}
\begin{tabular*}{\linewidth}{@{\extracolsep{\fill}}lccccccc@{}}
\toprule
\multicolumn{8}{@{}l}{\textbf{(a) Across foundation models: ChestAgentBench accuracy}}\\
& \multicolumn{3}{c}{Qwen family} & \multicolumn{4}{c}{GPT family}\\
\cmidrule(lr){2-4}\cmidrule(l){5-8}
Method & \shortstack{Qwen3.7-Flash} & \shortstack{Qwen3.7-Plus} & \shortstack{Qwen3.7-Max} & \shortstack{GPT-6 Luna} & \shortstack{GPT-6.1 Terra} & \shortstack{GPT-6.1 Sol} & \shortstack{GPT-6 Astra}\\
\midrule
Baseline & 73.75 & 73.42 &74.34&74.80&80.16& 82.54&82.38\\
Baseline+EGSE & \textbf{77.70} & \textbf{79.26} &\textbf{80.12}&\textbf{79.86}&\textbf{84.34}& \textbf{85.12}&\textbf{85.84}\\
\bottomrule
\end{tabular*}
\par\vspace{4pt}
\setlength{\tabcolsep}{2pt}
\begin{tabular*}{\linewidth}{@{\extracolsep{\fill}}lrrrrrrrrrr@{}}
\toprule
& \multicolumn{4}{c}{CheXbench} & \multicolumn{4}{c}{MIMIC-CXR} & \multicolumn{2}{c}{SLAKE-VQA}\\
\cmidrule(lr){2-5}\cmidrule(lr){6-9}\cmidrule(l){10-11}
Method & OpenI & Rad-Restruct & SLAKE & Overall & mF1-14 & MF1-14 & mF1-5 & MF1-5 & Accuracy & Recall\\
\midrule
\multicolumn{11}{@{}l}{\textbf{(b) Cross-dataset skill transfer: ChestAgentBench $\rightarrow$ other dataset}}\\
Baseline &59.7&66.9&89.4& 67.0	&77.3 & 33.6 & 69.8 & 49.3& 89.6 &91.2\\
Baseline+EGSE &\textbf{61.5}&\textbf{67.8}&\textbf{91.8}&\textbf{68.7}& \textbf{78.5}& \textbf{34.6} & \textbf{71.5}& \textbf{50.8} & \textbf{91.3} &\textbf{93.8}\\
\midrule
\multicolumn{11}{@{}l}{\textbf{(c) Within-dataset skill construction: construction split $\rightarrow$ test split (4:1)}}\\
Baseline &59.6&66.8&89.3& 66.9	&77.3 & 33.5 & 69.9 & 49.4& 89.7 &91.2\\
Baseline+EGSE & \textbf{63.4} & \textbf{68.5} & \textbf{92.6}& \textbf{70.2} & \textbf{78.9} & \textbf{35.4} & \textbf{71.9} & \textbf{51.7} &\textbf{91.9} & \textbf{94.5}\\
\bottomrule
\end{tabular*}\par\vspace{2pt}
{\raggedright\fontsize{7}{8}\selectfont All metrics are percentages (higher is better); Baseline is a ReAct-style agent that uses tools without skills, implemented with MedRAX~\cite{fallahpour2025medrax}; Baseline+EGSE augments this agent with EGSE. Table~\ref{tab:generalization}(a) constructs a separate library per FM. Table~\ref{tab:generalization}(b,c) uses Qwen3.7-Flash. In (b), the frozen ChestAgentBench library is evaluated on other full dataset. In (c), a library is constructed on four-fifths of each dataset and evaluated on the held-out fifth. The test sets therefore differ. 
\par}
\end{table*}

\noindent\textbf{Generalization across foundation models.}
Table~\ref{tab:generalization}(a) evaluates EGSE with a separately constructed library for each FM. EGSE improves ChestAgentBench accuracy across all seven backbones by 2.58--5.84\%. Gains extend from the default Qwen3.7-Flash (+3.95) to the stronger GPT backbones, including GPT-6.1 Sol (+2.58) and GPT-6 Astra (+3.46). These results demonstrate EGSE's generalizability across the evaluated foundation models.

\noindent\textbf{Cross-dataset skill transfer.}
Table~\ref{tab:generalization}(b) evaluates the frozen ChestAgentBench skill library on three external datasets. EGSE improves CheXbench pooled accuracy from 67.0\% to 68.7\% and SLAKE-VQA accuracy/recall from 89.6\%/91.2\% to 91.3\%/93.8\%. All four MIMIC-CXR F1 metrics also improve, by 1.0--1.7\%. The constructed skills thus generalize beyond the source dataset, including to report generation task.

\noindent\textbf{Within-dataset skill construction.}
Table~\ref{tab:generalization}(c) evaluates a separate skill library constructed within each dataset on its held-out test portion. EGSE improves CheXbench pooled accuracy by 3.3\%, SLAKE-VQA accuracy/recall by 2.2/3.3\%, and MIMIC-CXR F1 by 1.6--2.3\%. These results show that EGSE can construct skills that generalize to held-out cases across datasets.
\vspace{-5pt}

\subsection{Ablation Study}

\begin{figure*}[t]
\centering
\includegraphics[width=\linewidth]{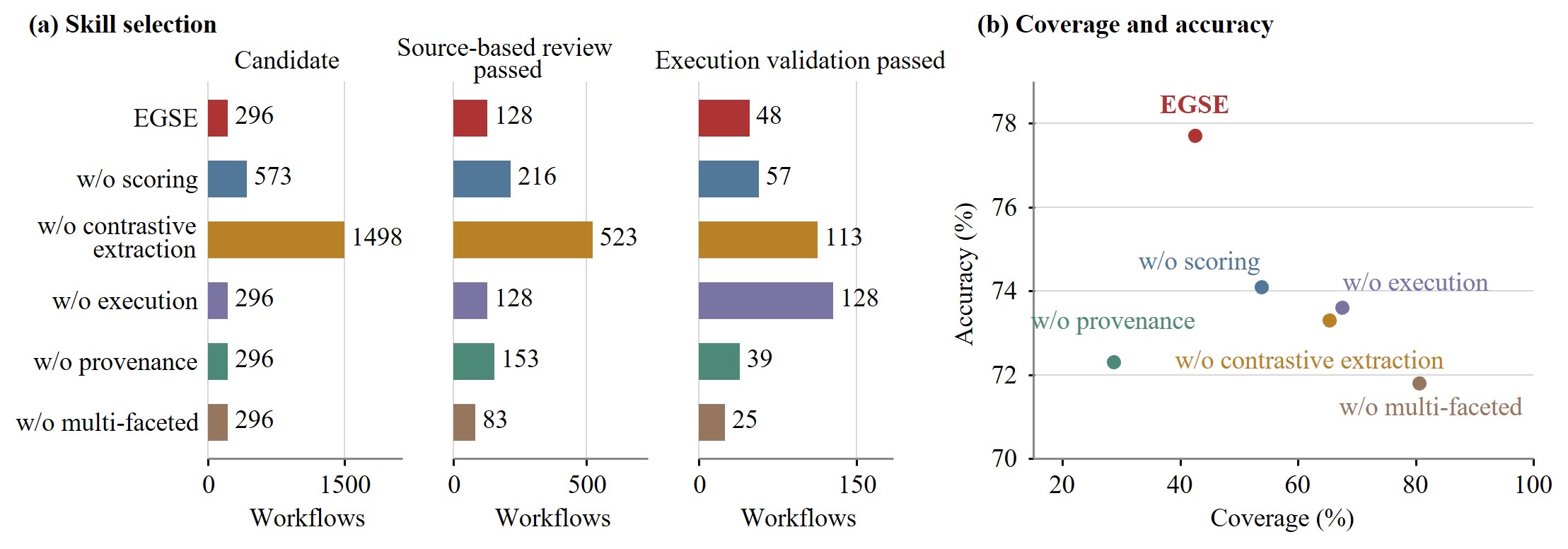}
\caption{{\textbf{Component ablations on workflow construction, coverage, and accuracy.} (a) Numbers of candidate workflows, source-based-review-passed workflows, and execution-validation-passed workflows, shown on separate scales. For the variant without execution validation, the last count denotes workflows accepted after source-based review. (b) Each point represents one setting. EGSE achieves the highest accuracy (77.7\%) with 48 execution-validation-passed workflows. Removing scoring, contrastive extraction, or execution validation increases accepted workflow counts and coverage but lowers accuracy. Removing provenance reduces coverage to 28.7\%, whereas removing multi-faceted organization raises it to 80.6\%; accuracy falls to 72.3\% and 71.8\%, respectively.}}
\label{fig:ablations}
\end{figure*}

We construct each ablated skill library from the same complete ChestAgentBench construction set of 1,498 cases and their grounded experiences. Candidate workflows comprise all workflows synthesized from the grounded experiences available under each setting, with at most one candidate per case. Figure~\ref{fig:ablations} reports the numbers of candidate workflows, workflows passing source-based review, and workflows passing execution validation. Coverage is the fraction of test cases receiving at least one selected workflow; accuracy includes all test cases. We ablate step scoring, contrastive extraction, execution validation, provenance constraints, and multi-faceted organization to assess usefulness, source support, and applicability. Here, the contrastive-extraction ablation jointly removes within-group step matching, common-experience extraction, and semantic subtraction. Appendix~\ref{supp:exp_controls} details the interventions and controlled settings.

\noindent\textbf{Usefulness of experience and updates.}
{Figure~\ref{fig:ablations} shows that removing step scoring or contrastive extraction increases candidate workflows from 296 to 573 and 1,498, respectively, but reduces accuracy from 77.7\% to 74.1\% and 73.3\%. Without step-quality signals or cross-trajectory comparison, incidental, redundant, and ineffective advice can more readily enter workflow construction. Removing execution validation accepts 128 workflows after source-based review, compared with 48 execution-validation-passed workflows in full EGSE, but lowers accuracy to 73.6\%. Together, these results suggest that step scoring and contrastive extraction help identify useful experience, while execution validation helps screen out ineffective library updates. }

\noindent\textbf{Source support of synthesized workflows.}
{Removing provenance increases source-based-review-passed workflows to 153, yet only 39 workflows pass execution validation, with coverage and accuracy falling to 28.7\% and 72.3\%. An independent LLM-as-a-Judge audit against the original construction dialogues finds source support for all 48 execution-validation-passed workflows of EGSE, compared with 13 of 39 (33.3\%) without provenance (Appendix~\ref{supp:workflow_support_audit}). These results show that execution success alone does not establish source support and suggest that provenance constraints help keep synthesized workflows grounded in the source dialogues.}

\noindent\textbf{Applicability of skill reuse.}
{Removing multi-faceted organization reduces execution-validation-passed workflows to 25, while coverage rises to 80.6\% and accuracy falls to 71.8\%. This pattern suggests that an unfaceted library permits broader reuse across cases whose requirements differ, without providing equally useful guidance. Full EGSE achieves higher accuracy with 42.5\% coverage. These results suggest that multi-faceted organization helps match workflows to the needs of the current case, rather than simply increasing how often workflows are used.}

\section{Conclusion}
\label{sec:conclusion}
We present EGSE, an evidence-grounded skill evolution framework for turning historical medical-agent dialogues into reusable guidance for new cases. By distilling conditional workflows from successful and failed rollouts, preserving their links to source evidence, and organizing them in a multi-faceted skill library, EGSE enables agents to accumulate and reuse procedural knowledge without updating model parameters. Experiments show improved reasoning accuracy with fewer tool calls and consistent gains across seven foundation models. Ablation studies suggest that step scoring, contrastive extraction, and execution validation help preserve useful experience in distilled skills, while evidence provenance supports grounded workflow synthesis and multi-faceted labels improve the relevance of injected skill. These findings highlight the value of grounding skills in historical evidence and matching their reuse to the procedural needs of each case.

\clearpage
\captionsetup[figure]{skip=\EGSEOriginalCaptionSkip}
\captionsetup[table]{skip=\EGSEOriginalCaptionSkip}
\setlength{\textfloatsep}{\EGSEOriginalTextFloatSep}
\setlength{\dbltextfloatsep}{\EGSEOriginalDblTextFloatSep}
\setlength{\floatsep}{\EGSEOriginalFloatSep}
\setlength{\dblfloatsep}{\EGSEOriginalDblFloatSep}
\setlength{\intextsep}{\EGSEOriginalInTextSep}
\section*{LLM Usage Statement}

The authors used GPT-6.1 Sol to assist with language polishing of author-written text, including grammar and spelling checks, to improve clarity and readability for non-native English writers. The tool also assisted in editing and refining author-produced code and implementation, as well as author-developed experimental design, execution, and analysis of results. Its role in these activities was limited to assistance with author-produced material. The full manuscript or major portions of it were not submitted to an LLM provider for reviewer-style feedback. The authors take full responsibility for all submitted content.

{
    \small
    \bibliographystyle{ieeenat_fullname}
    \bibliography{main}
}


\clearpage
\maketitlesupplementary
\appendix

This supplementary material begins with two qualitative case studies illustrating workflow reuse in Appendix~\ref{supp:case_studies}. Appendix~\ref{supp:prompts} then specifies the prompt interfaces for experience extraction, source-grounded workflow construction and update, and inference-time selection. Appendix~\ref{supp:exp_protocol} details the evaluation protocol, generalization settings, ablation controls, and independent workflow source-support audit. Appendix~\ref{supp:mqd} defines the facet taxonomy and input-only routing rules.

\noindent\textbf{Connection to the motivating questions.}
For usefulness, see the extraction prompts in Appendix~\ref{supp:prompt_extraction}, the update review and execution-validation details in Appendix~\ref{supp:prompts}, and the ablation controls in Appendix~\ref{supp:exp_controls}. For source support, see the construction prompts in Appendix~\ref{supp:prompt_construction} and the independent audit in Appendix~\ref{supp:workflow_support_audit}. For applicability, see the facet rules in Appendix~\ref{supp:mqd}, inference-time selection in Appendix~\ref{supp:prompt_inference}, and the qualitative case studies in Appendix~\ref{supp:case_studies}. The case studies illustrate workflow use, while the subsequent appendices organize implementation and evaluation details by method stage.

\section{Qualitative Case Studies}
\label{supp:case_studies}
\begingroup
\definecolor{caseblue}{RGB}{34,78,115}
\definecolor{casered}{RGB}{175,35,40}
\definecolor{casegreen}{RGB}{25,112,66}
\newcommand{\diff}[1]{\begingroup\setlength{\fboxsep}{1pt}\colorbox{yellow!35}{\strut #1}\endgroup}
\newcommand{\bad}[1]{\textcolor{casered}{#1}}
\newcommand{\good}[1]{\textcolor{casegreen}{#1}}
\newcommand{\mqdlabels}{\par\smallskip{\sffamily\footnotesize\textbf{M/Q/D:} \texttt{M\_XRAY / Q\_IDENTIFICATION / D\_12}\par X-ray / identification / respiratory-system topic.\par}\smallskip}
\newcommand{\reason}[1]{\par{\sffamily\footnotesize\color{black!70}\textit{Analysis:} #1\par}}
\newcommand{\casecomment}[1]{\par\smallskip{\sffamily\small\itshape\color{caseblue}\textbf{Comment:} #1\par}}
\newcommand{\llmstep}[2]{\par\smallskip\textcolor{caseblue}{\textbf{LLM #1}}\quad #2\par}
\newcommand{\toolstep}[2]{\par\smallskip\textcolor{violet}{\textbf{Tools #1}}\quad #2\par}
\tcbset{caseouter/.style={colback=white,colframe=caseblue,arc=3mm,boxrule=0.8pt,left=3mm,right=3mm,top=0.5mm,bottom=0.5mm,colbacktitle=caseblue,coltitle=white,fonttitle=\bfseries}}
\tcbset{injected/.style={colback=blue!3,colframe=caseblue!45,arc=1.5mm,boxrule=0.5pt,boxsep=0pt,left=1mm,right=1mm,top=1mm,bottom=1mm,before skip=3pt,after skip=4pt}}
\noindent Yellow marks key contrasts; \bad{red: error-path content}; \good{green: correct-path content}.\par\smallskip

Shaded boxes summarize injected guidance; LLM/Tools entries summarize execution. Gray Analysis notes and medical notes are author annotations, not agent inputs. A skill node groups workflows by M/Q/D; each workflow supplies a name, an applicability condition, and checks that connect observations to decisions. Classification assigns broad image labels; phrase grounding returns image regions for a requested phrase. Neither tool establishes a disease diagnosis.
\par\medskip
\noindent\textbf{\hyperlink{case-one}{Case 1}: reading the medical comparison.}  Reticulonodular means a net-like pattern mixed with small spots. Fibrosis means lung scarring; honeycombing means clustered cyst-like air spaces, a specific structural change rather than a synonym for scarring. Sarcoidosis is an inflammatory disease that can enlarge lymph nodes at the lung roots (the hila). Here, rejecting F for unverified honeycombing does not by itself prove B or its lymph-node finding.\par\medskip
\noindent\textbf{\hyperlink{case-two}{Case 2}: reading the medical comparison.}  Opacities are areas that look whiter than normal lung; consolidation describes denser air-space opacity. Perihilar means near the central lung roots; peripheral means near the outer lung edges. Organizing pneumonia is an inflammatory repair process, whereas BALT lymphoma is a slow-growing B-cell cancer involving lung-associated lymphoid tissue. Both can produce similar opacities: central location alone cannot distinguish them.\par\medskip
\noindent\textbf{How to read the analyses.} In \hyperlink{case-one}{Case 1}, ask whether a broad label (Fibrosis) verifies the specific feature stated in an answer (honeycombing). In \hyperlink{case-two}{Case 2}, ask whether a targeted location query supplies information used in the final option comparison. In both cases, distinguish the agent's interpretation from independently verified findings: a reference-matching answer alone does not establish that every medical inference is sound.
\par\medskip
\noindent\textbf{Medical background sources.} Sarcoidosis: \href{https://doi.org/10.1164/rccm.202002-0251ST}{ATS diagnostic guideline}. Overlapping appearances of organizing pneumonia and pulmonary lymphoma: \href{https://doi.org/10.1183/13993003.00158-2025}{ERS/ATS interstitial-pneumonia statement} and \href{https://doi.org/10.1038/s41598-019-45144-9}{53-patient pulmonary MALT lymphoma imaging study}. These sources explain terminology and diagnostic limits; they do not re-validate these individual images.
\begin{figure*}[!tp]
\begin{tcolorbox}[caseouter,title={Case 1: SkeMex vs EGSE --- identical tool feedback}]
\hypertarget{case-one}{}
\label{supp:case_same_feedback}

\noindent\begin{minipage}[c]{0.18\linewidth}
\centering
\includegraphics[width=\linewidth]{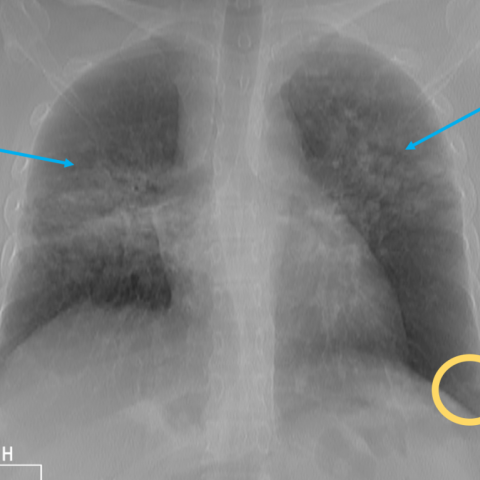}
\end{minipage}\hfill
\begin{minipage}[c]{0.78\linewidth}
\textbf{Input.} A 52-year-old woman with exertional dyspnea and dry cough for one month. Which diagnosis fits the reticulonodular chest X-ray?

\smallskip
\textbf{Relevant alternatives.} Sarcoidosis with bilateral hilar lymphadenopathy (B); pulmonary fibrosis with honeycombing (F).

\smallskip
\textbf{Ground-truth answer: B.}\quad Case ID: 13579.

\smallskip
\textit{The arrows and circle are present in the original input image.}
\end{minipage}

\smallskip
\noindent\colorbox{black!6}{\parbox{\dimexpr\linewidth-2\fboxsep\relax}{%

\textbf{Shared comparison conditions.} One classifier call per agent; identical full outputs. Selected scores: Lung Opacity 0.8611, Mass 0.5630, Nodule 0.5448, and Fibrosis 0.5205. These are label scores, not probabilities of the answer options. }}

\smallskip
\noindent\begin{minipage}[t]{0.48\linewidth}
\textbf{SkeMex}\par\smallskip\hrule\smallskip
\begin{tcolorbox}[injected]
\textbf{Injected skill.} \textit{Converge on Concordant Imaging and Classifier Evidence.}

\smallskip
\textit{Applies to:} closed-ended chest X-ray diagnosis; initial vision and classification suggest the same abnormality category.

\smallskip
\textit{Core-action excerpts:}\par
1. Call the classifier once to check the initial category.\par
2. Treat agreement as ``\diff{\bad{sufficient corroboration}}''.
\reason{LLM 2 treats Fibrosis 0.5205 as support for its initial diagnosis.}
3. Prefer the matching main diagnosis ``\bad{even if an associated imaging descriptor cannot be confidently resolved}''.
\reason{LLM 2 accepts F despite unconfirmed honeycombing.}
4. ``\diff{\bad{Stop evidence acquisition}} once this agreement is established''; seek more evidence only for direct contradiction.
\reason{LLM 2 stops after classification; honeycombing remains unchecked.}

\smallskip
\end{tcolorbox}
\llmstep{1}{Interprets reticulonodular changes as fibrosis; acknowledges difficulty identifying honeycombing, yet already favors \bad{F}. Requests classification.}
\toolstep{1}{\textbf{Classifier} $\rightarrow$ identical scores shown above.}
\llmstep{2}{Uses Fibrosis 0.5205 to reinforce the initial fibrosis interpretation. \diff{\bad{Honeycombing is unconfirmed}}, yet accepts F, whose main diagnosis matches that broad category. No further tool is requested.}\reason{Option F explicitly includes honeycombing. A broad Fibrosis score does not verify that feature; this is not a claim that all fibrosis requires honeycombing.}
\smallskip
\textbf{Final:} \bad{F: fibrosis with honeycombing. \textbf{Incorrect.}}\par
\textbf{2 LLM turns / 1 tool call.}

\end{minipage}\hfill
\begin{minipage}[t]{0.48\linewidth}
\textbf{EGSE}\par\smallskip\hrule\smallskip
\begin{tcolorbox}[injected]
\mqdlabels
\textbf{Injected skill.} Two workflows: \textit{Resolve PCP Pattern vs Classifier Noise} and \textit{Override Misleading Report in Cavitary Nodule Diagnosis}.

\smallskip
\textit{Check excerpts:} ``Inspect visual features for bilateral symmetry and fine reticular/linear markings in \diff{\good{perihilar regions}}''.
\reason{These checks correspond to LLM 1's visual comparison, before classification.}
``Confirm if nodules are multiple and bilateral/perihilar rather than solitary or upper-lobe predominant''.
\par\smallskip
``\diff{\good{unresolved imaging}} must leave the distinction open''.
\reason{The workflow discourages treating an uncertain feature as established.}
\smallskip
\end{tcolorbox}
\llmstep{1}{Following the skill's texture and distribution checks, identifies \good{fine reticulonodular markings with a diffuse/perihilar distribution}. This pattern favors \good{sarcoidosis (B)} over fibrosis with honeycombing (F). Requests classification.}
\reason{The skill turns texture and distribution into explicit criteria for favoring B over F.}
\toolstep{1}{\textbf{Classifier} $\rightarrow$ identical scores shown above.}
\llmstep{2}{Fibrosis 0.52 \diff{\good{does not confirm honeycombing}} specified in F. The observed reticulonodular pattern and perihilar distribution better support B. Selects \good{B: sarcoidosis}.}
\reason{The skill's uncertainty check prevents a broad Fibrosis label from being treated as confirmation of F, preserving the feature-based choice of B.}
\smallskip
\good{\textbf{Final: B: sarcoidosis. Correct.}}\par
\textbf{2 LLM turns / 1 tool call.}
\end{minipage}

\smallskip
\noindent\textbf{Summary.} \good{EGSE uses the skill's texture and distribution checks to favor B, then keeps the broad classifier label from overriding that feature-based comparison.}
\end{tcolorbox}
\end{figure*}
\begin{figure*}[!tp]
\begin{tcolorbox}[caseouter,title={Case 2: EGSE library expansion --- targeted evidence acquisition}]
\hypertarget{case-two}{}
\label{supp:case_targeted_query}

\noindent\begin{minipage}[c]{0.16\linewidth}
\centering
\includegraphics[width=\linewidth]{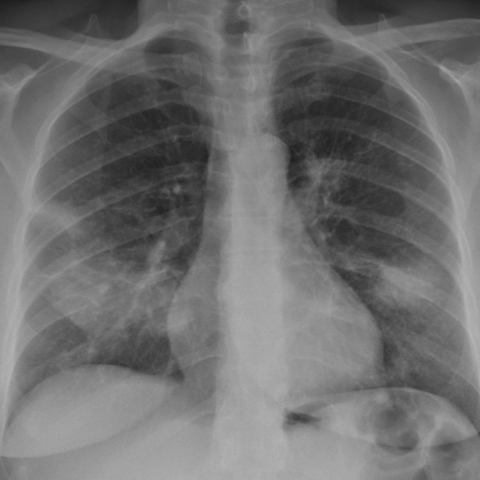}
\end{minipage}\hfill
\begin{minipage}[c]{0.80\linewidth}
\textbf{Input.} Select the diagnosis that best explains the pulmonary opacities on the chest X-ray.

\smallskip
\textbf{Relevant alternatives.} Organizing pneumonia (B); low-grade B-cell lymphoma of bronchus-associated lymphoid tissue, or BALT lymphoma (D). 

\smallskip
\textbf{Ground-truth answer: D.}\quad Case ID: 7484.

\smallskip
\textbf{Comparison.} Earlier versus expanded EGSE library; existing workflows are unchanged and four workflows are added across the library.
\end{minipage}

\smallskip
\noindent\colorbox{black!6}{\parbox{\dimexpr\linewidth-2\fboxsep\relax}{%

\textbf{Same initial classifier output.} Mass 0.8206, Lung Opacity 0.7289, and Nodule 0.6124. Calls within each batch below were requested in the same LLM turn. In this repeat, the earlier-library agent received no workflow; the final-library guidance is reproduced below.}}

\smallskip
\noindent\begin{minipage}[t]{0.48\linewidth}
\textbf{Earlier EGSE library}\par\smallskip\hrule\smallskip
\begin{tcolorbox}[injected]
\mqdlabels
    \textbf{Injected skill.} No workflow is present in the recorded injected context. The agent therefore uses its own reasoning and tools; this record alone does not identify whether retrieval or selection excluded the workflows.
\end{tcolorbox}
\llmstep{1}{Sees bilateral patchy/nodular opacities. Requests \diff{\bad{broad image characterization}}.}
\reason{The query does not isolate the distribution used later to distinguish B and D.}
\toolstep{1--3}{\textbf{Classifier + report + expert.}\par
Classifier: shared scores above. Report: no focal consolidation, multiple nodules. Expert: ``No significant findings''.}
\llmstep{2}{Notes conflicting descriptions and tool limitations. Requests more detail on distribution and morphology.}
\toolstep{4--6}{\textbf{Segmentation + grounding + VQA.}\par
Grounding query: \diff{\bad{``pulmonary nodules''}}.\par
VQA describes diffuse/miliary nodules.}
\reason{A general lesion query leaves the central-versus-peripheral comparison unresolved.}
\llmstep{3}{Acknowledges nonspecific findings, but \bad{selects organizing pneumonia from the patchy/nodular appearance}.}
\reason{This broad appearance does not explain why B should be preferred over D; the answer disagrees with the reference.}
\smallskip
\textbf{Final:} \bad{B: organizing pneumonia. \textbf{Incorrect.}}\par
\textbf{3 LLM turns / 6 tool calls.}

\end{minipage}\hfill
\begin{minipage}[t]{0.48\linewidth}
\textbf{Final EGSE library}\par\smallskip\hrule\smallskip
\begin{tcolorbox}[injected]
\small
\mqdlabels
\textbf{Injected Skill.} One workflow: \textit{Resolve PCP Pattern vs Classifier Noise}.\par
\par\smallskip
\textbf{Condition.} Bilateral interstitial/reticular opacities conflict with high mass/nodule/consolidation classifier scores.
\par\smallskip
\textit{Check excerpts:}\par
\textbf{1.} Inspect bilateral symmetry and fine reticular/linear markings in \diff{\good{perihilar regions}}. Prioritize specific imaging patterns over broad classifier labels.
\par\smallskip
\textbf{2.} Ground \diff{\good{perihilar abnormalities}} to check their spatial distribution and use it to compare the options.
\reason{The skill turns distribution assessment into an explicit localization target, providing spatial evidence for the subsequent comparison.}
\end{tcolorbox}
\llmstep{1}{Following the skill's distribution check, notes bilateral patchy opacities with a \good{central/perihilar distribution} and initially favors D. Requests classification and grounding of ``perihilar consolidation'' to check whether the abnormal regions match this distribution.}
\toolstep{1--2}{\textbf{Classifier + phrase grounding.}\par
Classifier: shared scores above.\par
Grounding query: \diff{\good{``perihilar consolidation''}}.\par
Return: \good{two regions covering bilateral central lung fields}.}
\reason{The agent turns the skill's perihilar check into a spatially targeted query, directly checking the distribution used to compare the options.}
\llmstep{2}{The returned regions match the visually assessed \good{bilateral central/perihilar distribution}. Combining the patchy opacities with their distribution, favors \good{BALT lymphoma (D)} over organizing pneumonia (B). Selects D.}
\reason{The agent uses localization in the B-versus-D comparison, explicitly linking its final choice to the queried spatial feature rather than mass or nodule labels alone.}
\smallskip
\good{\textbf{Final: D: BALT lymphoma. Correct.}}\par
\textbf{2 LLM turns / 2 tool calls.}
\end{minipage}

\smallskip
\noindent\textbf{Summary.} \good{EGSE turns the skill's distribution check into targeted localization, then combines the returned regions with visual assessment to compare the options and select D.}
\smallskip
\end{tcolorbox}
\end{figure*}
\endgroup

\section{Prompt Interfaces and Construction Details}
\label{supp:prompts}

This appendix specifies the prompt interfaces and construction procedures underlying experience extraction, source-supported workflow updates, and applicability assessment in Section~\ref{sec:method}.

\noindent\textbf{Prompt roles and information access.}
EGSE uses nine core prompts. The scoring, common-extraction, positive-experience, synthesis, relation, merge, and review prompts query $f_{\mathrm{ext}}$ during library construction. Labeling and selection query $f_\theta$ during both construction-time validation and test inference. These are separate stateless requests to the same foundation model. Among the nine core prompts, only step scoring receives the ground-truth answer directly. Common extraction receives the question and annotated step pairs; synthesis receives only extracted experiences. Task-specific outcome evaluation also uses ground-truth answers during construction, including the report judge in Appendix~\ref{supp:dataset_specific}. Test-time labeling and selection receive neither ground-truth answers nor historical outcome labels.

\noindent\textbf{Notation and record identifiers.}
Notation follows Section~\ref{sec:method}. The prompt listings retain their literal field names. In particular, their tool-call component \texttt{T} corresponds to $\mathcal{A}_l$ in Eq.~\eqref{eq:react}, and \texttt{steps} stores workflow checks. Record identifiers are lookup keys rather than additional mathematical variables: \texttt{X012.H+1.h3} identifies the third step of the first successful dialogue for case \texttt{X012}; \texttt{X012.C+(3,2)} identifies a selected positive pair containing steps 3 and 2; \texttt{X012.u+1} and \texttt{X012.u-1} identify retained positive and negative experiences; and \texttt{w3.s1} identifies a workflow check.

\noindent\textbf{Source-reference handling.}
Each extracted item stores supporting pair identifiers in \texttt{support\_refs}. Each synthesized check stores cited experience identifiers in \texttt{experience\_refs}, implementing $\mathcal{R}(s)$ in Eq.~\eqref{eq:provenance}. These records link checks to experiences and then to their original step pairs through $\mathcal{M}(u)$. During merging, \texttt{from\_steps} identifies existing checks whose references must be inherited, while \texttt{experience\_refs} identifies new supporting experiences. References are stored separately from check text and remain available for review; the agent receives only workflow names, conditions, and checks at inference time. The program checks identifier validity, while source-based review assesses whether the cited experiences support the content.

\noindent\textbf{Output checks and repair budgets.}
Malformed outputs and mechanically detectable constraint violations, such as unknown reference identifiers or exceeded length limits, trigger another request with the detected errors. The repair budgets are two attempts for synthesis, one for relation judgement, three for merging, and one for the review output format. A workflow that fails source-based review may be regenerated at most twice using the reported violations, as specified in Section~\ref{sec:update}. These repair requests are distinct from execution validation: a reviewed proposal is tested three times on each validation case, and any failed outcome prevents committing the change. The previous library is retained after rejection. Prompts are included from the project text files without changing their instructions; line wrapping and typographic normalization affect presentation only. The labeling listing abbreviates the full taxonomy as \texttt{label\_definitions}, which is supplied from Appendix~\ref{supp:mqd}.

\subsection{Experience Extraction}
\label{supp:prompt_extraction}

\subsubsection{Step Scoring Prompt \texorpdfstring{$P_{\mathrm{score}}$}{P-score}}
\label{supp:prompt_score}

As shown in Figure~\ref{fig:prompt_score}, $P_{\mathrm{score}}$ receives $X$, its ground-truth answer $y$, and the full dialogue $H$. It retrospectively evaluates each step in the context of the full dialogue and returns independent reasoning and tool-use scores, $\alpha_l$ and $\beta_l$, for every step, each with a one-sentence rationale (Eq.~\eqref{eq:score}). Both scores and rationales are attached to source steps before common extraction.

\egseprompt{STEP\_SCORING\_PROMPT}{prompts/p_score.tex}{Step scoring with independent reasoning and tool-use assessments.\label{fig:prompt_score}}

\subsubsection{Common Experience Prompt \texorpdfstring{$P_{\mathrm{com}}$}{P-com}}
\label{supp:prompt_common}

As shown in Figure~\ref{fig:prompt_com}, $P_{\mathrm{com}}$ receives $Q$ and the annotated pair sets $\widetilde{\mathcal{M}}^{\pm}$, which contain source identifiers, compared components, and both source steps' reasoning and tool-use scores and assessment rationales. Positive mode extracts $\mathcal{V}^+$ from $\widetilde{\mathcal{M}}^+$; negative mode extracts $\mathcal{U}^-$ from $\widetilde{\mathcal{M}}^-$ (Eq.~\eqref{eq:common}). Reasoning requires the corresponding reasoning scores to have the required sign on both sides; tool-use content independently requires the tool-use scores to do so. Zero scores and mismatched signs do not qualify. Eligibility alone is insufficient: the source steps and rationales must support shared advice. With reasoning-only pairs, a tool call on one side does not establish shared tool use.

The prompt requires each output item to contain \texttt{condition}, \texttt{statement}, \texttt{kind}, \texttt{evidence\_notes}, \texttt{limitations}, and nonempty \texttt{support\_refs}, and to return an empty list when no item qualifies. Scores and rationales guide assessment of the source content. The program assigns experience identifiers.

\egseprompt{COMMON\_EXPERIENCE\_PROMPT}{prompts/p_common.tex}{Score-guided extraction of shared recommendations or avoidance advice.\label{fig:prompt_com}}

\subsubsection{Positive Experience Prompt \texorpdfstring{$P_{\mathrm{pos}}$}{P-pos}}
\label{supp:prompt_pos}

As shown in Figure~\ref{fig:prompt_pos}, $P_{\mathrm{pos}}$ receives $\mathcal{V}^+$ and $\mathcal{U}^-$ and returns $\mathcal{U}^+$ (Eq.~\eqref{eq:positive}). It compares behaviors and applicability conditions, removing recommendations targeted by avoidance advice under overlapping conditions. Uncertain conflicts are dropped conservatively. A partial remainder must be independently supported by its original positive sources, and cannot borrow negative references as positive support. The unchanged $\mathcal{U}^-$ and retained $\mathcal{U}^+$ form $\mathcal{U}$ for synthesis. This filter does not verify negative advice beyond its recorded support.

\egseprompt{POSITIVE\_EXPERIENCE\_PROMPT}{prompts/p_pos.tex}{Conservative filtering of positive candidates using negative experience.\label{fig:prompt_pos}}

\subsection{Source-Grounded Workflow Construction and Update}
\label{supp:prompt_construction}

\subsubsection{Labeling Prompt \texorpdfstring{$P_{\mathrm{label}}$}{P-label}}
\label{supp:prompt_label}

As shown in Figure~\ref{fig:prompt_label}, $P_{\mathrm{label}}$ receives $X$ and the definitions of all 51 labels in Tables~\ref{tab:mqd_m}--\ref{tab:mqd_d}. It applies Appendix~\ref{supp:mqd_rules} and returns exactly one label per facet in the order M, Q, D, forming $\lambda$ in Eq.~\eqref{eq:mqd}. The same labeling procedure is used for construction cases and new cases.

\egseprompt{LABELING\_PROMPT}{prompts/p_label.tex}{Input-only facet labeling for skill-node routing.\label{fig:prompt_label}}

\subsubsection{Workflow Synthesis Prompt \texorpdfstring{$P_{\mathrm{syn}}$}{P-syn}}
\label{supp:prompt_syn}

As shown in Figure~\ref{fig:prompt_syn}, $P_{\mathrm{syn}}$ receives only $\mathcal{U}$ from one historical case. The prompt requests a candidate $w'$ with a name, condition, and one to four checks, preferably two or three. It limits the name to 60 characters, the condition to 150, and each check text to 180. The check-count limit applies within each workflow. Each check must cite at least one supplied experience and preserve its conditions, limitations, and role as a recommendation or avoidance lesson.

The prompt instructs the model to omit historical answers, option identifiers, image paths, and ground-truth language from every field, exclude specific disease names from names and conditions, and condition diagnoses inside checks on confirmed findings. The program assigns check identifiers and verifies reference identifiers; the subsequent source-based review assesses compliance with the content requirements.

\egseprompt{WORKFLOW\_SYNTHESIS\_PROMPT}{prompts/p_syn.tex}{Source-constrained synthesis of a transferable conditional workflow.\label{fig:prompt_syn}}

\subsubsection{Relation Prompt \texorpdfstring{$P_{\mathrm{rel}}$}{P-rel}}
\label{supp:prompt_rel}

As shown in Figure~\ref{fig:prompt_rel}, $P_{\mathrm{rel}}$ receives $w'$ and the workflows in $\mathcal{W}_\lambda$. It returns one relation per existing workflow, and a proposed condition change for each related pair. A duplicate requires coverage of the checks, their decision use, and the applicability condition. A supported new condition is considered an extension for merging rather than a reason to discard a proposal. The program skips a candidate if any pair is duplicate; otherwise, it considers related pairs, or proposes a separate addition when none is related.

For merging, the program chooses the related target requiring the smallest condition change: \texttt{none}, then \texttt{conditional}, then \texttt{generalize}. Ties follow the prompt's closeness rank. The field \texttt{scope\_change} stores this decision. A value of \texttt{none} preserves the condition while allowing supported refinement of checks; it does not mean that the proposal is a duplicate.

\egseprompt{RELATION\_PROMPT}{prompts/p_rel.tex}{Comparison of candidate and existing workflows by checks and condition coverage.\label{fig:prompt_rel}}

\subsubsection{Merge Prompt \texorpdfstring{$P_{\mathrm{merge}}$}{P-merge}}
\label{supp:prompt_merge}

As shown in Figure~\ref{fig:prompt_merge}, $P_{\mathrm{merge}}$ receives the selected existing workflow, $w'$, the experiences cited by $w'$, the condition-change decision, and a length budget of 1600 characters for the formatted workflow. The prompt requests one merged workflow with one to four checks, preserving narrower advice as conditional branches and requiring source support for a broader condition. It instructs the model to meet the budget by compressing wording while retaining checks and their decision use.

For each merged check, the program combines the references inherited through \texttt{from\_steps} with the new references in \texttt{experience\_refs}. The prompt permits rewording or compression of existing content to rely on inherited references alone, but requires a new experience reference for a new check or substantive content change. At least one reference list must be nonempty. Source-based review assesses whether these sources support the merged checks.

\egseprompt{MERGE\_PROMPT}{prompts/p_merge.tex}{Conditional merging with inherited and newly cited experience references.\label{fig:prompt_merge}}

\subsubsection{Review Prompt \texorpdfstring{$P_{\mathrm{rev}}$}{P-rev}}
\label{supp:prompt_rev}

As shown in Figure~\ref{fig:prompt_rev}, $P_{\mathrm{rev}}$ receives an added or merged workflow and its cited atomic experiences. It checks the four requirements in Section~\ref{sec:update}: checks must explain their effect on the decision; conclusions must follow actual check results; the entry condition must not assume an unresolved result; and all substantive content must be source-supported. A workflow passes only when no violation is reported. Otherwise, synthesis or merging is repeated with the violations as feedback, within the regeneration budget.

This construction-time review uses cited experiences. It is separate from the independent source-support audit in Appendix~\ref{supp:workflow_support_audit}, which evaluates workflows against original construction dialogues. Neither review nor that audit establishes independently verified clinical correctness.

\egseprompt{REVIEW\_PROMPT}{prompts/p_rev.tex}{Source-based review before execution validation.\label{fig:prompt_rev}}

\subsection{Inference-Time Workflow Selection}
\label{supp:prompt_inference}

\subsubsection{Selection Prompt \texorpdfstring{$P_{\mathrm{sel}}$}{P-sel}}
\label{supp:prompt_sel}

As shown in Figure~\ref{fig:prompt_sel}, $P_{\mathrm{sel}}$ receives $X$ and the names and conditions of workflows in $\mathcal{C}$ from Eq.~\eqref{eq:retrieval}, but no checks or source references. It returns at most two workflow identifiers with brief reasons, or an empty selection when no condition applies. Selected workflows should address distinct evidence tasks, and applicability is judged from the input without first guessing a diagnosis. The program looks up and formats their names, conditions, and checks as $\Gamma$ in Eq.~\eqref{eq:selection}. An empty result supplies no workflow guidance. The same procedure is used during execution validation and frozen-library test inference.

\egseprompt{WORKFLOW\_SELECTION\_PROMPT}{prompts/p_sel.tex}{Image-aware selection of applicable retrieved workflows.\label{fig:prompt_sel}}

\clearpage
\makeatletter\setlength{\@fptop}{0pt}\setlength{\@dblfptop}{0pt}\setlength{\@dblfpsep}{12pt}\makeatother

\clearpage
\begingroup
\raggedbottom
\section{{Evaluation Protocol and Audit}}
\label{supp:exp_protocol}

{This appendix details the evaluation settings used in Section~\ref{sec:experiments}. Cross-dataset skill transfer evaluates a frozen source library on each full target evaluation pool, whereas within-dataset skill construction uses a grouped construction/test split. The split audit below reports metadata-level checks; it does not establish image-content deduplication.

}

\subsection{Dataset Partition and Access Rules}

\noindent\textbf{Reproducible source partition.}
The source metadata is from ChestAgentBench~\cite{fallahpour2025medrax}.
The split manifest records source case IDs, question IDs, and partition assignments. In the main text, each question-based evaluation instance is termed a case; the raw case IDs remain source grouping keys. {The 2,500 clinical cases correspond to 609 patients. The construction/development/test partitions contain 1,498/503/499 cases from 365/122/122 patients, respectively; all cases from one patient remain in one partition.} Its accompanying summary records the metadata and manifest SHA-256 hashes. A fixed search over 2,000 seeded case permutations minimizes relative deviations from the desired question and task-label counts; answers and model predictions do not enter this objective. The seven task annotations overlap, so their counts must not be added to obtain the number of questions. This is a custom partition for supervised library construction; published full-benchmark scores are not directly comparable to its held-out results.

\noindent\textbf{Remaining independence checks.}
The metadata audit finds no cross-partition shared image paths or image-source URLs. Exact and perceptual image hashes still require verification for duplicate images under different identifiers; linked cases must remain in one partition if duplicates are found.  Any resulting changes require a versioned manifest and reevaluation. The released metadata includes one question without a listed image, assigned to the test partition. {The primary score retains} all 499 released test questions; a secondary score on the 498 image-bearing questions must be reported consistently for all methods. Tool or API failures must not silently remove examples from either denominator.

\noindent\textbf{External datasets and metrics.}
The external pool definitions follow Section 4.1 (p.~5) of the supplied EATS manuscript: {CheXbench includes 115 Rad-Restruct VQA, 238 SLAKE VQA, and 380 OpenI image--text reasoning cases}, SLAKE-VQA includes 114 closed-ended English chest X-ray cases, and MIMIC-CXR includes 3,858 single-image studies. {These full evaluation pools are used for cross-dataset skill transfer in Table~\ref{tab:generalization}(b), whereas Table~\ref{tab:generalization}(c) uses only the test portions of the within-dataset splits.} CheXbench reports separate accuracies for OpenI, Rad-Restruct, and SLAKE. Overall is pooled accuracy: the sum of correct predictions across these three source subsets divided by their total evaluated case count, rather than an unweighted average of subset accuracies. Report each subset denominator after the grouped split. For SLAKE-VQA, freeze answer normalization and word tokenization before computing exact-match accuracy and reference-word recall. For MIMIC-CXR, freeze the CheXbert version and treatment of uncertain and missing labels, and aggregate by study for micro-/macro-F1 over 14 and 5 findings. {CheXbert~\cite{smit2020combining} extracts CheXpert labels~\cite{irvin2019chexpert}; the five-finding subset comprises cardiomegaly, edema, consolidation, atelectasis, and pleural effusion. All F1 metrics are percentages.} Report-generation F1 must not be described as accuracy.

Detect overlap between SLAKE-derived CheXbench items and the SLAKE pool, and between ChestAgentBench construction data and each target pool, using source identifiers and image-content checks. Shared images must not occur on opposite sides of construction/test partitions, including across the two VQA datasets. A target item that overlaps source construction must be removed from both cross-dataset skill transfer and within-dataset skill construction test manifests. Audit neural-tool training overlap separately from skill-library data access.

\subsection{Backbone Validation}
\label{supp:exp_models}

{The default backbone and extraction model is Qwen3.7-Flash, with temperature $0.2$, top-$p$ $0.95$, rollout cap $R=8$, reasoning-step budget $L=7$, and 768-dimensional bge-base-en-v1.5 embeddings.

}

\noindent\textbf{Hardware.}
All experiments are conducted on two NVIDIA RTX Pro 6000 GPUs (96GB VRAM each).

The FM panel comprises seven variants spanning different capacity tiers in two families: Qwen3.7-flash, Qwen3.7-Plus, and Qwen3.7-Max; GPT-6 Luna, GPT-6.1 Terra, GPT-6.1 Sol, and GPT-6 Astra. Qwen3.7-flash is the default backbone. For each model, record the actual endpoint, dated snapshot where available, returned model identifier, execution dates, and decoding/reasoning configuration used in the experiments. The display names alone do not identify an exact API snapshot; model-specific settings must be documented rather than assumed identical across providers.

\noindent\textbf{Backbone-specific construction.}
For each model, generate a history on the same construction partition and use that model for both $f_{\theta}$ and $f_{\mathrm{ext}}$. Compare Direct FM, MedRAX, MedRAX+SkeMeX, MedRAX+Trace2Skill, and MedRAX+EGSE under shared tools, data access, and inference limits. Document the adapters for the two published skill methods and retain their original construction and usage mechanisms; these are MedRAX-based adaptations rather than copied published scores. Reuse the same recorded history across compatible memory-method comparisons within a backbone. Report the number of cases yielding two correct and two incorrect rollouts, extraction failures, library size, and total construction cost. Different models may have different eligible-case sets; publish those sets and include a shared-case sensitivity analysis before attributing differences to model generality.

{\noindent\textbf{Interpretation of the FM comparison.}
Table~\ref{tab:generalization}(a) rebuilds the library for each backbone; it evaluates the construction procedure across FMs rather than transfer of one fixed library between FMs.

}

\subsection{Within-Dataset Skill Construction with a 4:1 Split}
\label{supp:dataset_specific}
\label{tab:dataset_specific}

{
For Table~\ref{tab:generalization}(c), each external evaluation pool is divided approximately 4:1 into construction and test portions. Questions sharing an image or source record remain together for CheXbench and SLAKE-VQA, and MIMIC-CXR is grouped by patient. These are custom within-dataset splits rather than unchanged official test sets.

Three independent target libraries are constructed from the respective construction portions and frozen for evaluation on the held-out portions. Target test cases and labels are excluded from extraction, revision, validation, and configuration selection. This setting evaluates whether EGSE can construct useful skills within different datasets.

For Table~\ref{tab:generalization}(b), the ChestAgentBench library is reused unchanged on each full target evaluation pool, without target construction trajectories or library updates. Both settings use the default FM, neural tools, inference limits, and evaluation metrics. Baseline denotes MedRAX without skills, and Baseline+EGSE augments it with the corresponding library. Each within-setting comparison uses the same test cases, but Table~\ref{tab:generalization}(b) and Table~\ref{tab:generalization}(c) use different test sets and need not have identical Baseline scores. Their absolute scores cannot directly isolate the benefit of within-dataset skill construction over cross-dataset skill transfer.

}

\noindent\textbf{Report-generation construction.}
{For MIMIC-CXR construction, an LLM-as-a-Judge compares the generated report with the reference report and the question using three binary criteria: faithfulness, relevance, and coherence. Faithfulness requires preservation of the reference's clinically material findings, negation, and uncertainty without unsupported additions; relevance requires answering the requested reporting task; coherence requires an internally consistent and understandable report. A rollout is successful only if all three criteria pass. The same rule determines execution-validation acceptance for each of the three runs per validation case. These judgements define construction outcomes; the reported CheXbert F1 metrics remain unchanged. The proposed judge prompt is given below.}

\egseprompt{REPORT\_JUDGE\_PROMPT}{prompts/p_report_judge.tex}{{Proposed binary judge for report-generation construction and execution validation.}}

\subsection{Five Ablation Groups}
\label{supp:exp_controls}

{The primary ablation uses all 1,498 ChestAgentBench construction cases and the same recorded rollout pool for Full EGSE and five variants on the default backbone. Each setting synthesizes at most one candidate workflow per case from its available grounded experiences.} Hold the historical rollout pool, input order, FM, tool versions, workflow-length limit, repair caps, and inference-context cap fixed. Rebuild each library under the stated intervention. Use the same construction inputs, rather than stopping after an equal number of approved workflows. Actual computation and final library size are outcomes to report.

\noindent\textbf{Without ground-truth step scoring.}
{Remove step scores, their rationales, and sign-based eligibility rules. Extract common practices from successful dialogues as positive candidates and common evidenced problems from failed dialogues as negative experience, without replacing scores with another quality judge.} Retain the same recorded successful and failed rollouts, outcome labels, common extraction, semantic subtraction, provenance, and execution validation. This tests the contribution of intermediate scoring, not the removal of all ground-truth supervision.

\noindent\textbf{Without contrastive extraction.}
{Retain the same rollout pool, outcome annotations, step scores, and assessment rationales,} but extract advice independently from individual trajectories without the two-rollout common-support requirement or {subtraction of negative experience from positive candidates}. Feed this advice to the same synthesis stage under the same length budget. Provenance maps to the source steps actually used; it must not require absent paired-step fields. Do not force every failed-trajectory action to become avoidance advice. This jointly ablates within-group step matching, common-experience extraction, and semantic subtraction, testing the contrastive-extraction mechanism as a whole without identifying their individual effects. Paired rollouts remain recorded to control data and sampling cost.

\noindent\textbf{Without execution validation.}
Accept nonduplicate additions and merges after the same source-based review and repair procedure, omitting only the execution gate. Record submitted and accepted proposals, final library coverage, and test accuracy. Lower approval in Full EGSE is useful if accompanied by better final performance. This comparison does not require approval to increase with quality. If the paper makes the narrower claim that existing behavior is preserved, supplement final accuracy with matched pre/post update outcomes on a fixed set of earlier cases, recording correct-to-incorrect transitions and whether the edited workflow was selected. Such checks are diagnostics, not an additional ablation group or a guarantee against all regressions.

\noindent\textbf{Without provenance constraints.}
Use the same atomic experiences, but remove check-level source-link requirements and the source-link-based restrictions on synthesis and merging. Retain length limits, general semantic/content review, and execution validation. A missing reference field must not itself cause rejection in this variant. {Audit the execution-validation-passed workflows against the original dialogues of all cases used in their synthesis and updates, using the procedure and denominator in Appendix~\ref{supp:workflow_support_audit}.}

\noindent\textbf{Without multi-faceted organization.}
Place workflows in a shared, unfaceted library during construction and inference. Retain multiple conditional workflows, condition retrieval, image-aware selection, the same context cap, and all extraction and acceptance rules. This tests whether grouping by modality, question type, and disease helps workflows meet appropriate updates and be used in appropriate cases, relative to the same conditional machinery without facets. It does not separately isolate the selector or the number of workflows per node. Those distinctions should not be claimed from this single variant. Higher coverage is not necessarily better: relevant routing can reduce inappropriate reuse while increasing accuracy.

\subsection{Ablation Metrics and Figure Conventions}
\label{supp:exp_plots}

{Figure~\ref{fig:ablations}(a) reports numbers of candidate workflows, source-based-review-passed workflows, and execution-validation-passed workflows. Candidate workflows are all workflows synthesized from the case-level grounded experiences, before source-based review and execution validation. Counts refer to workflow proposals at each stage, including merged workflow proposals, rather than numbers of nonempty nodes or necessarily the final library size: a passing merge replaces an existing workflow. The three count axes use separate scales and start at zero. Without execution validation, the 128 source-based-review-passed workflows are accepted directly; this is the explicitly indicated exception to the execution-validation-passed count. Figure~\ref{fig:ablations}(b) uses one circular marker per setting, identified by its adjacent label. Marker shape has no additional statistical meaning.}

Coverage is the fraction of test cases receiving at least one selected workflow,
\begin{equation}
\mathrm{Coverage}=\frac{100}{N_{\mathrm{test}}}
\sum_{n=1}^{N_{\mathrm{test}}}
\mathbb{I}\!\left[|\rho(\mathcal{X}_n;\mathcal{S})|>0\right].
\end{equation}
It measures selected and injected guidance, not the proportion of populated taxonomy nodes. Accuracy uses the full test denominator, rather than only cases receiving guidance. The reported values come from one run per setting; no repeated-run variability, error bars, or confidence intervals are reported. The comparison does not establish statistical significance.

\subsection{{Workflow Source-Support Audit}}
\label{supp:workflow_support_audit}
{An independent LLM-as-a-Judge audits every execution-validation-passed workflow against the original dialogues of the cases used for its synthesis and all subsequent updates. Both settings use the same rubric and provide original dialogue content, rather than relying on experience summaries or reference identifiers. The judge is blinded to the method name and reported execution outcomes. Each workflow is supported only when its applicability condition and every substantive check, tool instruction, and decision branch are supported by the source dialogues. A recorded agent assertion or an observed action alone does not establish the validity of the advice; unsupported conclusions and unjustified generalizations fail. The judge cites source-step evidence for each assessment. Support rate is the number of supported workflows divided by the number of execution-validation-passed workflows. This audit measures dialogue-source support, not independently verified clinical correctness.}

\begin{table}[t]
\centering

\caption{{Source support among execution-validation-passed workflows.}}
\label{tab:workflow_support_audit}
\setlength{\tabcolsep}{3pt}
\begin{tabular*}{\linewidth}{@{\extracolsep{\fill}}lccc@{}}
\toprule
 & \shortstack{Execution validation\\passed workflows} & \shortstack{Supported\\workflows} & \shortstack{Support\\rate}\\
\midrule
EGSE & 48 & 48 & 100\%\\
w/o provenance & 39 & 13 & 33.3\%\\
\bottomrule
\end{tabular*}
\end{table}

\egseprompt{WORKFLOW\_SUPPORT\_JUDGE}{prompts/p_support_judge.tex}{{Judge rubric for auditing workflow support in original construction dialogues.}}

\subsection{Tool Analysis}
Step-level $\beta_i$ and call-level quality require separate denominators. No-tool steps do not enter a tool-step mean, and Direct FM has no tool quality score. Aggregate individual-call scores by tool alongside call counts and state whether an overall mean weights calls or questions. Use a common rubric and a method-blinded audit sample. Distinguish source support, tool-output correctness, and usefulness to the final decision; provenance or a higher model-assigned score alone does not establish clinical correctness.

\clearpage
\endgroup

\section{Multi-Faceted Label Taxonomy}
\label{supp:mqd}

\begin{table*}[!t]
\centering
\footnotesize
\caption{Imaging modality (M) labels.}
\label{tab:mqd_m}
\begin{tabular}{@{}>{\raggedright\arraybackslash}p{0.20\textwidth}>{\raggedright\arraybackslash}p{0.29\textwidth}>{\raggedright\arraybackslash}p{0.45\textwidth}@{}}
\toprule
Label ID & Name & Definition \\
\midrule
\texttt{M\_XRAY} & X-ray radiography & Projection X-ray imaging, including chest and skeletal radiographs; excludes CT. \\ \addlinespace[3pt]
\texttt{M\_CT} & Computed tomography & CT slices, reconstructions, and volumetric images, including contrast-enhanced CT. \\ \addlinespace[3pt]
\texttt{M\_MRI} & Magnetic resonance imaging & Various MRI sequences, slices, and reconstructions. \\ \addlinespace[3pt]
\texttt{M\_US} & Ultrasound & Includes Doppler ultrasound and echocardiography. \\ \addlinespace[3pt]
\texttt{M\_PET} & Positron emission tomography & Functional or metabolic PET images. \\ \addlinespace[3pt]
\texttt{M\_NUCLEAR\_OTHER} & SPECT and other nuclear imaging & Nuclear medicine images other than PET. \\ \addlinespace[3pt]
\texttt{M\_PATH\_MICRO} & Histopathology and microscopy & Histological, cytological, and medical microscopy images. \\ \addlinespace[3pt]
\texttt{M\_ENDOSCOPY} & Endoscopy & Endoscopic images, including gastrointestinal endoscopy and bronchoscopy. \\ \addlinespace[3pt]
\texttt{M\_CLINICAL\_PHOTO} & Clinical photography & Photographs of the body surface, skin, and wounds, including dermoscopic photographs; excludes endoscopic, fundus, microscopy, and OCT images. \\ \addlinespace[3pt]
\texttt{M\_FUNDUS} & Fundus imaging & Fundus photographs; excludes OCT. \\ \addlinespace[3pt]
\texttt{M\_OCT} & Optical coherence tomography & OCT cross-sectional images. \\ \addlinespace[3pt]
\texttt{M\_MULTIMODAL} & Multiple imaging modalities & At least two imaging modalities that require joint processing, including PET/CT fusion; multiple images of the same modality do not qualify. \\ \addlinespace[3pt]
\texttt{M\_UNKNOWN} & Unknown modality & Insufficient evidence to determine the modality, or a modality outside the categories above. \\
\bottomrule
\end{tabular}
\end{table*}

\begin{table*}[!t]
\centering
\footnotesize
\caption{Question task type (Q) labels.}
\label{tab:mqd_q}
\begin{tabular}{@{}>{\raggedright\arraybackslash}p{0.20\textwidth}>{\raggedright\arraybackslash}p{0.29\textwidth}>{\raggedright\arraybackslash}p{0.45\textwidth}@{}}
\toprule
Label ID & Name & Definition \\
\midrule
\texttt{Q\_PRESENCE} & Presence or verification & Determine whether a specified target is present. \\ \addlinespace[3pt]
\texttt{Q\_IDENTIFICATION} & Identification or classification & Determine identity or category, including the most likely diagnosis. \\ \addlinespace[3pt]
\texttt{Q\_LOCALIZATION} & Localization & Determine location, laterality, region, or spatial relationships. \\ \addlinespace[3pt]
\texttt{Q\_ATTRIBUTE} & Attribute assessment & Assess shape, margins, density, signal, or color; excludes location, count, measurement, and severity. \\ \addlinespace[3pt]
\texttt{Q\_COUNTING} & Counting & Determine the number of targets. \\ \addlinespace[3pt]
\texttt{Q\_MEASUREMENT\_GRADING} & Measurement or grading & Assess length, area, volume, size, severity, or grade. \\ \addlinespace[3pt]
\texttt{Q\_COMPARISON} & Comparison or change & Compare or assess changes across time points, images, objects, or regions. \\ \addlinespace[3pt]
\texttt{Q\_DESCRIPTION} & Comprehensive description & Provide overall findings or a report rather than a single local operation. \\ \addlinespace[3pt]
\texttt{Q\_UNKNOWN} & Unknown task type & The task is unclear, not covered by the categories, or contains multiple equally primary independent tasks. \\
\bottomrule
\end{tabular}
\end{table*}

\begin{table*}[!t]
\centering
\footnotesize
\caption{Disease category (D) labels. The scope summarizes the ICD-11 chapter description or, for chapters without one (\texttt{D\_04}, \texttt{D\_12}, \texttt{D\_13}, \texttt{D\_17}, \texttt{D\_25}, \texttt{D\_26}), the main blocks of the chapter~\cite{who2026icd11}. Routing notes state the rules of Appendix~\ref{supp:mqd_rules} that apply to the chapter.}
\label{tab:mqd_d}
\begin{tabular}{@{}>{\raggedright\arraybackslash}p{0.15\textwidth}>{\raggedright\arraybackslash}p{0.24\textwidth}>{\raggedright\arraybackslash}p{0.56\textwidth}@{}}
\toprule
Label ID & Chapter name & Scope and routing note \\
\midrule
\texttt{D\_01} & Certain infectious or parasitic diseases & Conditions caused by pathogenic organisms, such as bacteria, viruses, parasites, or fungi. \textit{Routing:} not every infection belongs here; pneumonia $\rightarrow$ \texttt{D\_12}, infections of the fetus or newborn $\rightarrow$ \texttt{D\_19}. \\ \addlinespace[3pt]
\texttt{D\_02} & Neoplasms & Abnormal or uncontrolled cellular proliferation that is not coordinated with the requirements for normal tissue growth, replacement, or repair; covers benign, in situ, malignant, and uncertain-behaviour neoplasms. \textit{Routing:} includes hematologic malignancies and lung cancer, regardless of the organ involved. \\ \addlinespace[3pt]
\texttt{D\_03} & Diseases of the blood or blood-forming organs & Diseases of the blood and of blood-forming organs, \eg, anaemias, coagulation defects, and diseases of the spleen. \textit{Routing:} hematologic malignancies $\rightarrow$ \texttt{D\_02}; immune-system diseases $\rightarrow$ \texttt{D\_04}. \\ \addlinespace[3pt]
\texttt{D\_04} & Diseases of the immune system & Immunodeficiencies, autoimmune and autoinflammatory disorders, allergic or hypersensitivity conditions, and diseases of the thymus. \textit{Routing:} not merged into blood diseases. \\ \addlinespace[3pt]
\texttt{D\_05} & Endocrine, nutritional or metabolic diseases & Endocrine diseases, nutritional disorders, and metabolic disorders. \\ \addlinespace[3pt]
\texttt{D\_06} & Mental, behavioural or neurodevelopmental disorders & Syndromes with clinically significant disturbance in cognition, emotional regulation, or behaviour that reflects dysfunction in psychological, biological, or developmental processes. \\ \addlinespace[3pt]
\texttt{D\_07} & Sleep-wake disorders & Difficulty initiating or maintaining sleep, excessive sleepiness, sleep-related breathing disorders, circadian rhythm disorders, sleep-related movement disorders, and parasomnias. \textit{Routing:} sleep-related breathing disorders belong here. \\ \addlinespace[3pt]
\texttt{D\_08} & Diseases of the nervous system & Conditions in or associated with the nervous system, \eg, epilepsy, movement disorders, and cerebrovascular diseases. \textit{Routing:} cerebrovascular diseases belong here; vascular keywords do not imply \texttt{D\_11}. \\ \addlinespace[3pt]
\texttt{D\_09} & Diseases of the visual system & Diseases of the eyes and ocular adnexa, the visual pathways, and the brain areas involved in vision. \\ \addlinespace[3pt]
\texttt{D\_10} & Diseases of the ear or mastoid process & Diseases of the external, middle, and inner ear and of the mastoid process, including disorders with hearing impairment. \\ \addlinespace[3pt]
\texttt{D\_11} & Diseases of the circulatory system & Diseases of the system that transports nutrients, gases, and other substances to and from body cells, \ie, the heart, arteries, veins, and lymphatic vessels. \textit{Routing:} vascular keywords do not override membership in other chapters. \\ \addlinespace[3pt]
\texttt{D\_12} & Diseases of the respiratory system & Diseases of the upper and lower respiratory tract, lung infections, interstitial lung diseases, pleural, diaphragmatic, or mediastinal disorders, and respiratory failure. \textit{Routing:} includes pneumonia; lung cancer $\rightarrow$ \texttt{D\_02}; not every chest-radiograph question belongs here. \\ \addlinespace[3pt]
\texttt{D\_13} & Diseases of the digestive system & Diseases of the digestive tract from the orofacial complex to the anal canal, and of the liver, biliary tract, pancreas, and peritoneum, including hernias. \\ \addlinespace[3pt]
\texttt{D\_14} & Diseases of the skin & Conditions affecting the epidermis and its appendages, associated mucous membranes, the dermis, the cutaneous vasculature, and the subcutaneous tissue. \\
\bottomrule
\end{tabular}
\end{table*}

\begin{table*}[!t]
\centering
\footnotesize
{\small Table~\ref{tab:mqd_d} (continued).}\par\vspace{6pt}
\begin{tabular}{@{}>{\raggedright\arraybackslash}p{0.15\textwidth}>{\raggedright\arraybackslash}p{0.24\textwidth}>{\raggedright\arraybackslash}p{0.56\textwidth}@{}}
\toprule
Label ID & Chapter name & Scope and routing note \\
\midrule
\texttt{D\_15} & Diseases of the musculoskeletal system or connective tissue & Diseases of the musculoskeletal system and connective tissue, \eg, arthropathies, spinal conditions, soft tissue disorders, and osteopathies. \textit{Routing:} traumatic fractures $\rightarrow$ \texttt{D\_22}. \\ \addlinespace[3pt]
\texttt{D\_16} & Diseases of the genitourinary system & Diseases characterised by pathological changes to the genitourinary system and the breast. \\ \addlinespace[3pt]
\texttt{D\_17} & Conditions related to sexual health & Sexual dysfunctions, sexual pain disorders, and gender incongruence. \\ \addlinespace[3pt]
\texttt{D\_18} & Pregnancy, childbirth or the puerperium & Conditions occurring from conception to delivery, during labour and delivery, and in the puerperium after delivery. \textit{Routing:} pregnancy alone does not assign a question on another topic. \\ \addlinespace[3pt]
\texttt{D\_19} & Certain conditions originating in the perinatal period & Conditions that originate in the perinatal period, even when death or morbidity occurs later. \textit{Routing:} classified by perinatal origin; the condition may present at a later age. \\ \addlinespace[3pt]
\texttt{D\_20} & Developmental anomalies & Conditions caused by the failure of a body site or body system to develop correctly, including multiple anomaly syndromes and chromosomal anomalies. \\ \addlinespace[3pt]
\texttt{D\_21} & Symptoms, signs or clinical findings, not elsewhere classified & Less well-defined symptoms, signs, and findings, \eg, when no more specific diagnosis is reached after investigation or the cause of a transient symptom is undetermined. \textit{Routing:} not a fallback for uncertain diagnoses. \\ \addlinespace[3pt]
\texttt{D\_22} & Injury, poisoning or certain other consequences of external causes & Physical or physiological bodily harm from interaction with energy or from a lack of vital elements, including poisoning and toxic effects of substances. \textit{Routing:} the injury itself, including traumatic fractures. \\ \addlinespace[3pt]
\texttt{D\_23} & External causes of morbidity or mortality & The external causes of injury, grouped by intent, \eg, unintentional causes, assault, and exposure to extreme forces of nature. \textit{Routing:} the cause rather than the injury itself. \\ \addlinespace[3pt]
\texttt{D\_24} & Factors influencing health status or contact with health services & Encounters with health services that are not driven by a current disease or injury, \eg, vaccination or organ donation, and circumstances that influence health status. \\ \addlinespace[3pt]
\texttt{D\_25} & Codes for special purposes & Codes for the provisional assignment of new diseases of uncertain aetiology and for emergency use. \textit{Routing:} must not be used as an unknown category. \\ \addlinespace[3pt]
\texttt{D\_26} & Supplementary chapter: traditional medicine conditions & Traditional medicine conditions, organized in Modules I and II. \textit{Routing:} traditional medicine classification only. \\ \addlinespace[3pt]
\texttt{D\_V} & Supplementary section for functioning assessment & Functioning profiles and scores of individuals, assessed with ICF-based instruments such as WHODAS 2.0 and the Model Disability Survey. \textit{Routing:} not arbitrary questions about organ function. \\ \addlinespace[3pt]
\texttt{D\_NOT\_APPLICABLE} & ICD chapter not applicable & Not an ICD-11 chapter: topics clearly outside the disease and health classification, \eg, purely normal anatomy or device localization. \\ \addlinespace[3pt]
\texttt{D\_UNKNOWN} & ICD chapter undetermined & Not an ICD-11 chapter: health-related topics whose chapter cannot be determined reliably from the input, or that span several chapters without a primary topic. \\
\bottomrule
\end{tabular}
\end{table*}

The facet tuple $\lambda$ introduced in Eq.~\eqref{eq:mqd} identifies a skill node. Tables~\ref{tab:mqd_m}, \ref{tab:mqd_q}, and~\ref{tab:mqd_d} specify its modality, task, and disease-topic labels. For example, (\texttt{M\_XRAY}, \texttt{Q\_IDENTIFICATION}, \texttt{D\_12}) denotes X-ray inputs, an identification task, and a question topic in the respiratory-system chapter. It does not assert a respiratory diagnosis or establish that every workflow in the node applies to the case. Conditions and image-aware selection provide the subsequent applicability checks in Section~\ref{sec:inference}. Label IDs are internal routing identifiers rather than ICD codes. Appendix~\ref{supp:mqd_rules} gives the labeling rules, and Appendix~\ref{supp:mqd_labels} describes the label tables.

\subsection{Labeling Rules}
\label{supp:mqd_rules}

The labels are assigned once per case by $f_{\theta}$ with the labeling prompt $P_{\mathrm{label}}$ (Appendix~\ref{supp:prompt_label}), which presents the case together with the definitions of all labels and returns one label per facet in the order M, Q, D. The labeler never accesses the ground truth, historical answers, tool outputs, or step scores. The three facets follow different rules.
\begin{itemize}
\item \textbf{M} describes the images actually provided, not the tools the agent intends to call. Reliable modality metadata take precedence over the image content, and a textual mention of a modality (\eg, ``CT'') does not imply that such an image is provided. Multiple images of one modality remain a single-modality input; \texttt{M\_MULTIMODAL} requires at least two modalities that must be processed jointly.
\item \textbf{Q} is determined by the operation that the question ultimately requests, not by the answer format, the tools, or intermediate steps. For example, asking whether a finding is on the left is localization, whereas asking whether it has enlarged since an earlier image is comparison. \texttt{Q\_UNKNOWN} covers questions with several equally primary independent tasks.
\item \textbf{D} is predicted from the question topic and the visible information, without first guessing the diagnosis. A question asking whether pneumonia is present belongs to \texttt{D\_12} even when the answer is no, a lung-cancer topic belongs to \texttt{D\_02}, and a generic question about chest-radiograph abnormalities does not automatically belong to \texttt{D\_12}. Official chapter membership takes precedence over organ, specialty, or keyword shortcuts. \texttt{D\_UNKNOWN} is used when membership is uncertain or several chapters are involved without a clearly primary topic, and ICD-11 extension codes are not used as routing chapters.
\end{itemize}

\subsection{Label Definitions}
\label{supp:mqd_labels}

Tables~\ref{tab:mqd_m}, \ref{tab:mqd_q}, and~\ref{tab:mqd_d} define all M, Q, and D labels, respectively. For each D label, Table~\ref{tab:mqd_d} gives the scope of the corresponding ICD-11 chapter and, where relevant, the routing rule for ambiguous topics.

\clearpage

\end{document}